\documentclass[letterpaper]{article} % DO NOT CHANGE THIS
\usepackage{aaai24}  % DO NOT CHANGE THIS
\usepackage{times}  % DO NOT CHANGE THIS
\usepackage{helvet}  % DO NOT CHANGE THIS
\usepackage{courier}  % DO NOT CHANGE THIS
\usepackage[hyphens]{url}  % DO NOT CHANGE THIS
\usepackage{graphicx} % DO NOT CHANGE THIS
\usepackage{natbib}  % DO NOT CHANGE THIS AND DO NOT ADD ANY OPTIONS TO IT
\usepackage{caption} % DO NOT CHANGE THIS AND DO NOT ADD ANY OPTIONS TO IT
\usepackage{algorithm}
\usepackage{algorithmic}
\usepackage{subcaption}
\usepackage{amsmath}
\usepackage{cleveref}
\usepackage{xcolor}

\makeatletter
\def\copyright@on{}
\makeatother

\usepackage{newfloat}
\usepackage{listings}
\DeclareCaptionStyle{ruled}{labelfont=normalfont,labelsep=colon,strut=off} % DO NOT CHANGE THIS
\floatstyle{ruled}
\newfloat{listing}{tb}{lst}{}
\floatname{listing}{Listing}
\title{VJT : A Video Transformer on Joint Tasks of 
Deblurring, Low-light Enhancement and Denoising}
\author {
    Yuxiang Hui\textsuperscript{\rm 1},
    Yang Liu\textsuperscript{\rm 2},
    Yaofang Liu\textsuperscript{\rm 3}
    Fan Jia\textsuperscript{\rm 1}\\
    Jinshan Pan\textsuperscript{\rm 4}
    Raymond Chan\textsuperscript{\rm 3}
    Tieyong Zeng\textsuperscript{\rm 1}
}
\affiliations {
    \textsuperscript{\rm 1}The Chinese University of Hong Kong, 
    \textsuperscript{\rm 2}National University of Defense Technology, \\
    \textsuperscript{\rm 3}City University of Hong Kong, 
    \textsuperscript{\rm 4}Nanjing University of Science and Technology\\

}

\usepackage{bibentry}
\begin{document}

\maketitle

\begin{abstract}
Video restoration task aims to recover high-quality videos from low-quality observations. This contains various important sub-tasks, such as video denoising, deblurring and low-light enhancement, since video often faces different types of degradation, such as blur, low light, and noise. Even worse, these kinds of degradation could happen simultaneously when taking videos in extreme environments. This poses significant challenges if one wants to remove these artifacts at the same time. In this paper, to the best of our knowledge, we are the first to propose an efficient end-to-end video transformer approach for the joint task of video deblurring, low-light enhancement, and denoising. This work builds a novel multi-tier transformer where each tier uses a different level of degraded video as a target to learn the features of video effectively. Moreover, we carefully design a new tier-to-tier feature fusion scheme to learn video features incrementally and accelerate the training process with a suitable adaptive weighting scheme. We also provide a new Multiscene-Lowlight-Blur-Noise (MLBN) dataset, which is generated according to the characteristics of the joint task based on the RealBlur dataset and YouTube videos to simulate realistic scenes as far as possible. We have conducted extensive experiments, compared with many previous state-of-the-art methods, to show the effectiveness of our approach clearly.
\end{abstract}

%% main text
\section{Introduction}
\label{sec:intro}

% Reason for why we need Joint Tasks
Image and video restoration, which aim
to provide high-quality videos, have been long-standing important tasks in computer vision. Most of the existing approaches to image and video restoration problems design models and datasets for a specific task, such as video deblurring \citep{wang2019edvr,zhong2020estrnn,Pan2020TSP,son2021recurrent,zhong2021towards,ji2022multi,liang2022vrt,fgst}, denoising \citep{maggioni2012video,arias2018video,chen2016deep,qi2022real,tassano2019dvdnet}, low light enhancement\citep{guo2016lime,fu2015probabilistic,lore2017llnet,lv2018mbllen,peng2022lve,zhang2021learning,xu2020learning,guo2020zerodce,triantafyllidou2020low}, deraining\citep{yue2021semi_rain,yang2019frame_rain,li2018recurrent_rain,wang2019spatial_rain,zhang2022enhanced_rain}. However, in the real world, many conditions that lead to low-quality video often occur simultaneously. For instance, we can have low light, blur, and noise from relatively long exposures at the same time when we take videos at night. This raises the urgent need to handle joint recovery tasks on images and videos. 
%and this is evidently difficult and interesting.

% Figure of our results, LEDNet and FGST
\begin{figure}
\setlength{\belowcaptionskip}{0.1cm}
\setlength{\abovecaptionskip}{0.1cm}
	\centering
	\begin{subfigure}{.23\textwidth}
		\centering
		\includegraphics[width=0.9\linewidth]{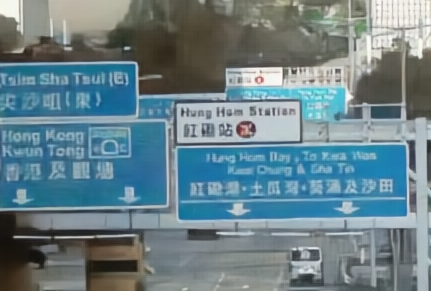}
		\caption{LEDNet}
		\label{fig:c1}
	\end{subfigure}\hspace{-3mm}
	\begin{subfigure}{.23\textwidth}
		\centering
		\includegraphics[width=0.9\linewidth]{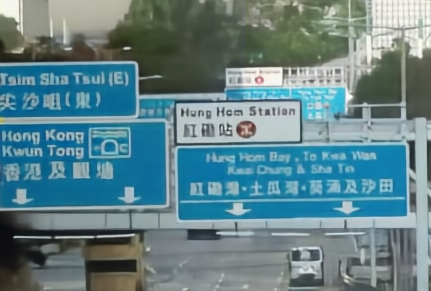}
		\caption{RVRT}
		\label{fig:c2}
	\end{subfigure}
 
	\begin{subfigure}{.23\textwidth}
		\centering
		\includegraphics[width=0.9\linewidth]{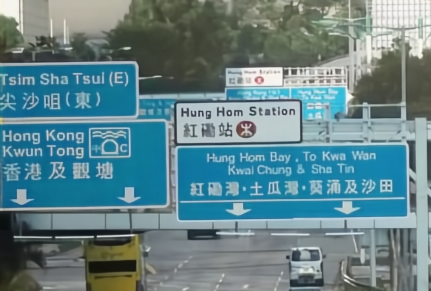}
		\caption{Our Method}
		\label{fig:c3}
    \end{subfigure}\hspace{-3mm}
    \begin{subfigure}{.23\textwidth}
		\centering
		\includegraphics[width=0.9\linewidth]{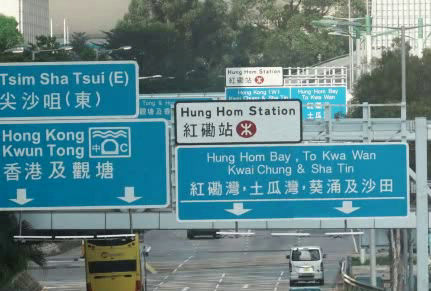}
		\caption{Ground Truth}
		\label{fig:c4}
	\end{subfigure}
	\caption{Restoration results on a part of a frame from a daytime outdoor scene of our dataset. Our method simultaneously removes blur, enhances low light and eliminates noise.  Compared to three recent methods retrained on our MLBN dataset, our method improves image quality in terms of object detail.}
	\label{fig:fig1}
\end{figure}

% Existing methods achieve good results on a single task
% For deblur, Low-light enhancement, denoise, show some methods.
Note that video deblurring, low-light enhancement, and denoising are among the most critical and challenging tasks in the field of video restoration. Many previous recurrent methods have proven effective in solving video recovery problems due to the sequential character of videos. In particular, several transformer models
\citep{RohitGirdhar2018VideoAT,liu2021video,liang2022vrt,liang2022recurrent,gberta_2021_ICML, arnab2021vivit} have performed exceptionally well for these tasks separately in recent years. 

% Explaining the disadvantages of putting different models in series.
 As existing deblurring methods, light enhancement and denoising methods can deal with each problem individually, one can simply combine these methods sequentially for the joint task. However, this direct combination may not achieve the best results for low quality images and videos suffered from blurring, low visibility and heavy noise simultaneously.
 The reason for this unsatisfactory result is clear. Indeed, a system processing low-quality videos to high-quality videos sequentially will lose some valuable information step by step and may revoke unknown noise and artifacts for the coming individual task. The joint task can better utilise all the valuable information for video restoration.
 
In this regard, it is natural for us to consider the joint task of video deblurring, low-light enhancement and denoising simultaneously to get the best restoration results. Immediately, we need to handle two critical issues: a suitable dataset for the joint task and a leading joint video restoration method. Our paper will address both issues effectively.

% Clarify our dataset and dataset generation process
Indeed, the first aim of our work is to provide a suitable dataset for the joint task of deblurring, lowlight enhancement and denoising. This Multi-scenes Lowlight-Blur-Noise (MLBN) dataset contains 195 scenes, covering indoor, nighttime outdoor and daytime street scenes, which are often involved in video recording. We propose a data generation progress to obtain realistic low-light blurry videos with noise. For the blurring part, we use the method in \citep{SeungjunNah2016DeepMC} to gain the corresponding blurry and clear frames from the high frame rate videos. As for the low-light part, we design an inverse LIME-based algorithm, considering the Retinex model LIME\citep{guo2016lime}. In addition, we have made different degrees of degradation for both the illumination and reflectance components, which will give more realistic results than simply using the inverse LIME algorithm. 
Finally, we employ the CycleISP algorithm to synthesise noise.

% Clarify our method and its advantages
Based on our progress in data generation for the MLBN dataset, videos of different stages of degradation are produced. 
Inspired by this, our second aim is to propose a leading joint video restoration method. Indeed, we design a multi-tier video transformer framework whereby each tier uses ground truth videos with different degradation levels as targets. A feature fusion method is used between tiers to transfer the features learned in the previous tier to the next one. This allows the network to step up the learning of features to three subtasks, resulting in a better reconstruction.
In addition, we employ an adaptive weight scheme to cope with the multiple loss functions of the joint task. It allows the different loss functions to reach the same energy level, thus making the training process more efficient. This accelerates the training and enables higher results at each stage of the progressive joint task.

% summary of our contribution
In summary, the main contributions of our paper are as follows:
\begin{itemize}

\item We propose the VJT, a {\bf Multi-tier Video Transformer} that can get differentiated features from three progressive tasks. We use the {\bf Feature Fusion} between tiers to make the structure more efficient. We also use an {\bf Adaptive Weight Scheme} to balance the magnitude of different losses to speed up training and achieve better results.

% \item We are the first to propose an end-to-end Video Transformer on the joint task of video deblurring, low-light enhancement and denoising.

\item We are pioneering researchers to address the integrated challenges of {\bf Joint Video Restoration} by using our VJT.

\item We propose a new data generation progress that balances the visibility, information density and noise intensity well in low-light videos to approximate real scenes closely enough. We have generated a new {\bf Multi-scene Lowlight-Blur-Noise (MLBN) Dataset} for our joint task.

\end{itemize}

% Figure of Framework
\begin{figure*}[htp]
    \centering
    \includegraphics[width=1\linewidth]{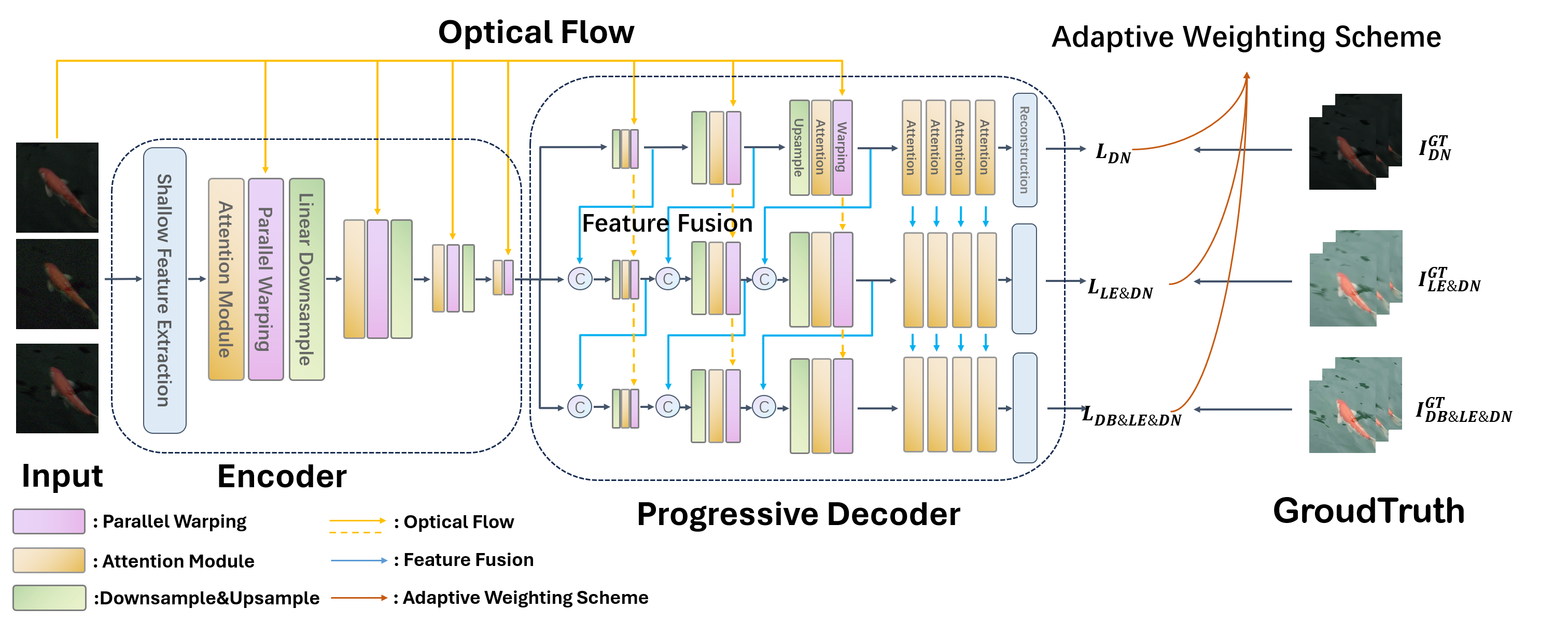}
    \caption{An illustration of the proposed VJT. It contains a single tier Encoder and a Multi-tier Decoder. There are multiple scales Attention-Warping modules in both the encoder and decoder. A shallow feature extraction module begins, while four attention and a reconstruction module are at the end.
    Each tier obtains a video with a different level of restoration. Feature fusion modules between tiers can transfer features for progressive joint tasks. The skip connections have been omitted for clarity.}
    \label{fig:arch}
\end{figure*}

\section{Related Work}
%\subsection{Video Transformer}

\subsection{Video Transformer.} 
Transformer-based models \citep{vaswani2017attention,dosovitskiy2020vit,carion2020end,liu2021swin,liu2021swinv2,liu2021video,xie2021moby} have shown effective results for various vision tasks.  The Swin-Transformer proposed by Liu \citep{liu2021swin,liu2021swinv2,liu2021video,xie2021moby} has performed very well in multiple areas such as objection detection, semantic segmentation, and action classification.
Compared to image tasks, video transformers need to consider one more temporal attention, and several works are excellent on video tasks\citep{RohitGirdhar2018VideoAT,liu2021video,liang2022vrt,liang2022recurrent,gberta_2021_ICML, arnab2021vivit}.
Liang \citep{liang2022vrt} proposed a Video Restoration Transformer that combines space attention and temporal attention based on Swin Transformer. Then they suggested RVRT \citep{liang2022recurrent}, which uses a globally recurrent framework and locally temporal attention module.
Bertasius \citep{gberta_2021_ICML} compared different paradigms that unify temporal and spatial attention.

%\subsection{Video Restoration}
\subsection{Deblurring.} 
Traditional methods \citep{YunpengLi2010GeneratingSP,TaeHyunKim2015GeneralizedVD} are based on image and video priors and assumptions. In recent years, as computing power has increased, deep learning methods have begun to be applied on a large scale \citep{JonasWulff2014ModelingBV,su2017deep,gong2017motion,hyun2017online,wang2019edvr,zhong2020estrnn,Pan2020TSP,ji2022multi,son2021recurrent,zhong2021towards}. Most of them use CNN, RNN-based methods. Ji \citep{ji2022multi} designed a multi-scale bidirectional recurrent method using a memory-based feature aggregation. In the last two years, with the popularity of Transformer architecture, there are several articles\citep{fgst,liang2022vrt,liang2022recurrent,zhang2022spatio,zhong2020estrnn} using a transformer for video deblurring. 
Wang \citep{wang2019edvr} proposed a pyramid, cascading and deformable convolution module for alignment and a TSA module for feature fusion. They were early adopter of the attention model for deblurring.
Zhang \citep{zhang2022spatio} proposed a spatio-temporal deformable attention module for video deblurring.
Lin \citep{fgst} customised a flow-guided sparse window-based multi-head self-attention module for video deblurring. 

\subsection{Low-light Enhancement.} 
Traditional low light enhancement methods include Retinex-based \citep{guo2016lime,fu2015probabilistic,hao2019low,park2017low} and histogram-based methods \citep{ibrahim2007brightness,abdullah2007dynamic}. The method based on the Retinex model is used more often; it generally decomposes the low-light image into illumination and reflection components by some regularisation or a prior condition and then enhances the illumination component to obtain the normal light image. LIME is a representative of Retinex models, which proposes two algorithms for accelerating the calculation of illumination maps. 
The deep learning-based approach also achieves strong results in image and video low-light enhancement \citep{lore2017llnet,lv2018mbllen,peng2022lve,zhang2021learning,xu2020learning,guo2020zerodce,triantafyllidou2020low}. \citep{lv2018mbllen} extracted features through a feature extraction module, an enhancement module, and a fusion module.
There is also a group of models that combine traditional Retinex methods with deep networks \citep{wei2018deep,yang2021sparse,li2018lightennet,wang2019underexposed,zhang2019kindling}. 
\citep{wei2018deep} used a decom-net to divide the image into reflective and illuminated components and an enhancement-net for low light enhancement.

% Figure of Feature Fusion and Attention Module
\begin{figure*}[ht]
\centering
    \begin{subfigure}{.5\textwidth}
        \centering
        \includegraphics[width=0.8\linewidth]{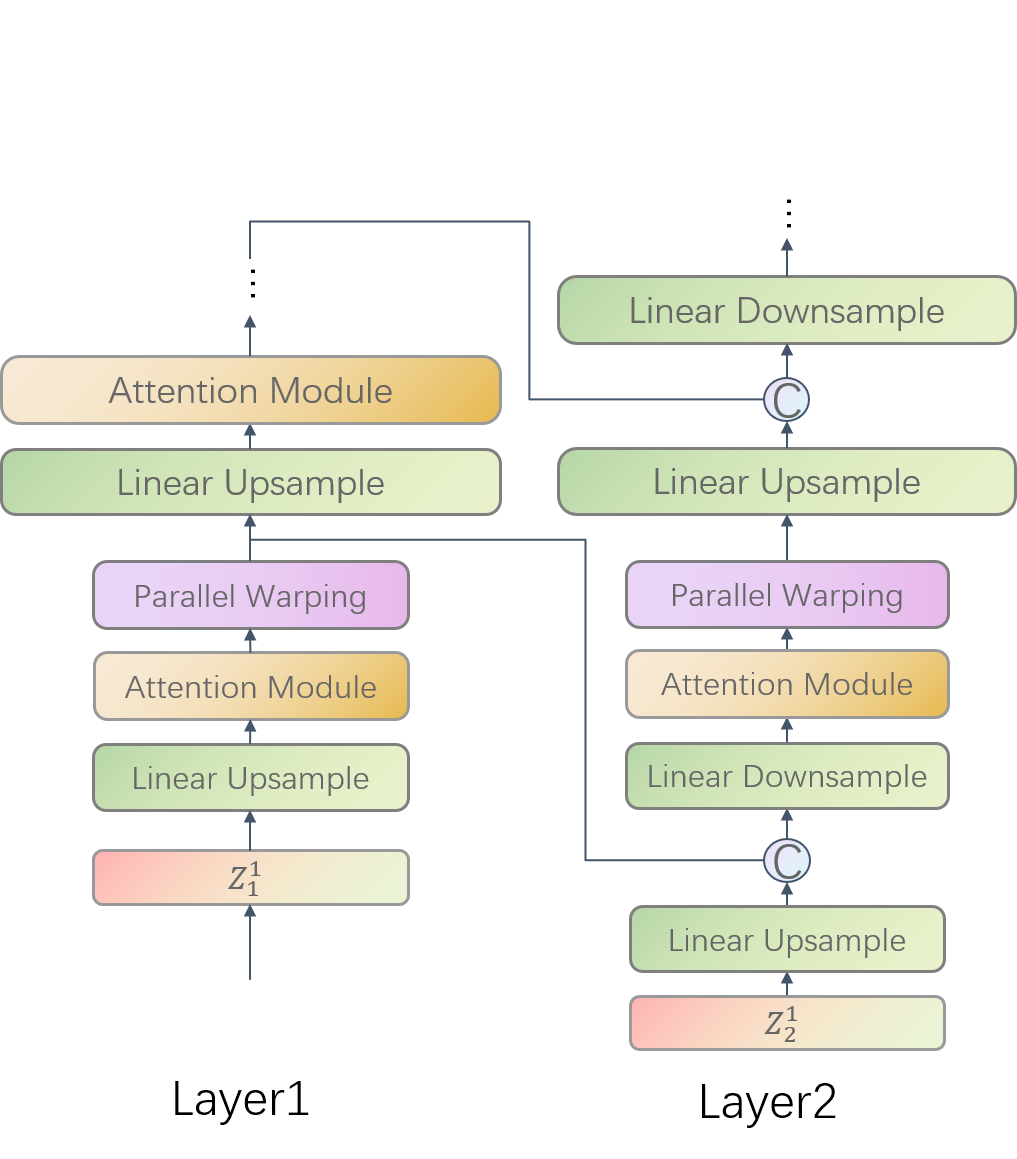}
        \caption{Feature Fusion}
        \label{fig:fusion}
    \end{subfigure}%
    \begin{subfigure}{.5\textwidth}
        \centering
        \includegraphics[width=0.8\linewidth]{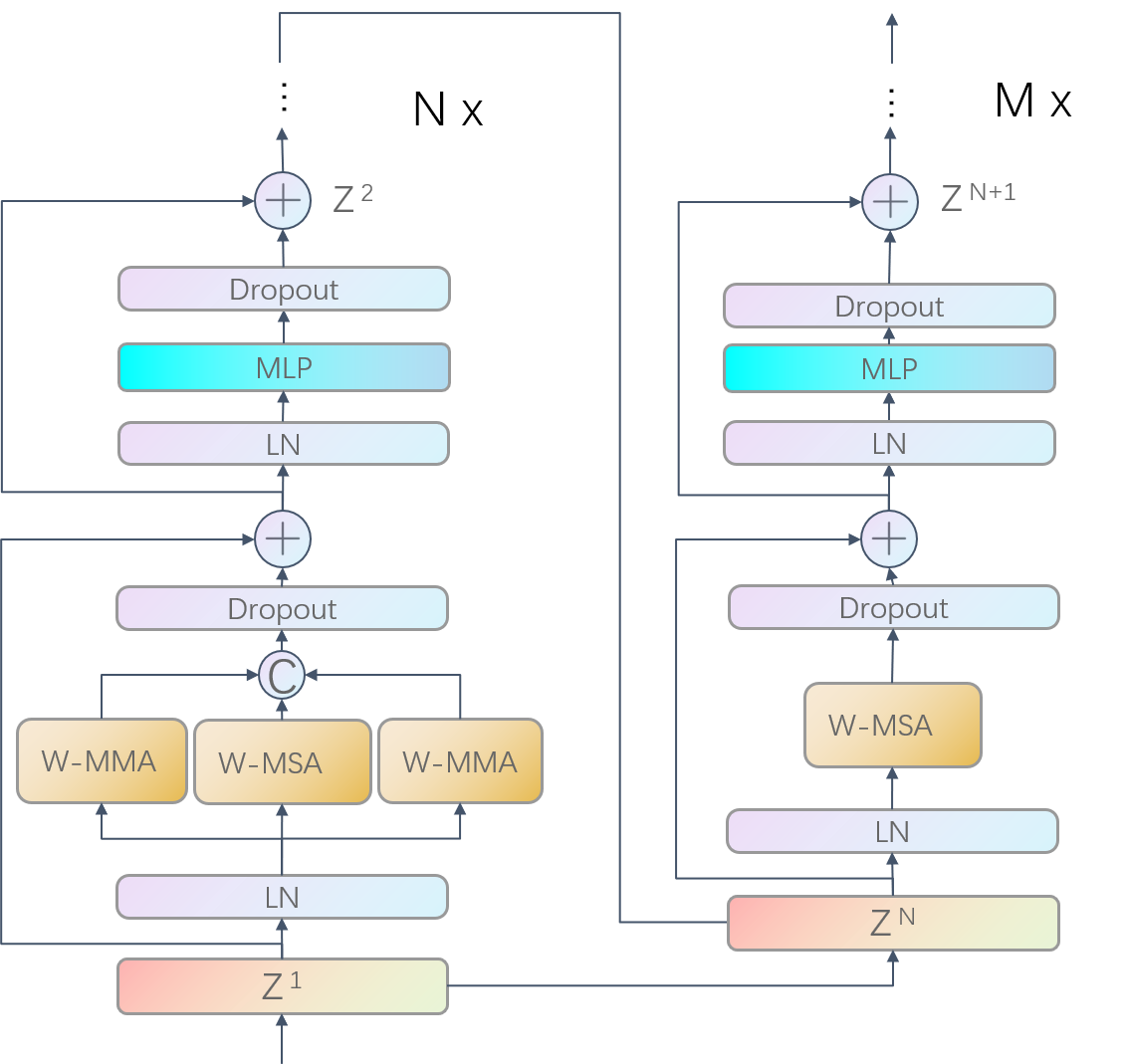}
        \caption{Attention Module}
        \label{fig:attention}
    \end{subfigure}%
    \caption{Illustrations for Feature Fusion between tiers and Attention Module in VJT. (a) shows the feature fusion between tier1 and tier2 at the first Attention-Warping module of the Decoder. (b) illustrates the attention module, which consists of $N$ sub-parts containing W-MMA and $M$ sub-parts with only W-MSA.}
\end{figure*}

\subsection{Denoising.} 

%Prior traditional approaches like \citep{maggioni2012video,arias2018video} are based on BM3D. Among the deep learning approaches for video denoising in recent years\citep{chen2016deep,qi2022real,tassano2019dvdnet}, many recurrent frameworks based on CNN have been applied to capture information in the time domain. \citep{qi2022real} introduced a novel bidirectional buffer block as the core module of their bidirectional streaming video denoising (BSVD) framework.

Prior traditional approaches to video denoising, such as those by Maggioni et al. \citep{maggioni2012video} and Arias et al. \citep{arias2018video}, are based on BM3D. Recent deep learning approaches, like those by Chen et al. \citep{chen2016deep} and Qi et al. \citep{qi2022real}, have utilized CNNs for capturing temporal information. Qi et al. introduced a novel bidirectional buffer block in their BSVD framework \citep{qi2022real}.

Recent advancements include the work of Wang et al. \citep{wang2023gen}, who extended text-driven generative models to long video generation with temporal co-denoising. Chan et al. \citep{chan2022generalization} adapted the BasicVSR++ framework for video denoising, demonstrating the effectiveness of long-term propagation and alignment. Yue et al. \citep{yue2020supervised} introduced a novel dataset for dynamic scene video denoising with their RViDeNet. Tassano et al. \citep{tassano2020fastdvdnet} presented FastDVDnet, notable for its real-time denoising capabilities. Finally, Claus and van Gemert \citep{claus2019videnn} introduced ViDeNN, a CNN for blind video denoising, emphasizing the importance of specialized training data.

\subsection{Joint Tasks.} 
There is some previous work on images for joint tasks of two. Zhou \citep{zhou2022lednet} proposed a CNN-based network for the joint task of low-light enhancement and deblurring on images, which uses Low-light enhancement encoder and deblurring decoder with filter skip connection. Zhao \citep{zhao2022d2hnet} designed a two-stage method D2HNet for denoising and deblurring on images. Both methods produce their datasets for the joint tasks on image restoration.

For unifying the three tasks, Xu \citep{xu2022deep} proposed a kernel prediction network for the joint Video Super Resolution, low-light enhancement and denoising. They used both static and horizontal motion datasets.

\section{Our Method} % Polished BY GPT4
\label{sec:method}

% \subsection{Overall Framework}
% % add motivation
% We regard the joint task of deblurring, low-light enhancement and denoising as a non-blind video restoration problem. We use synthetic datasets' characteristics and different degrees of "ground truth" under the progressive relationship to design a 3-tier decoder structure. Between tiers, we created a feature fusion module to transfer relatively shallow and single features to learning deep features. And to make the three loss functions at the same level, we designed an adaptive weight scheme for better training.

% We denote ${I^{LQ}}$ as the input blur, low-light and noise (Low quality) frames, and ${I_i^{R}}$ as the output of each of the three tiers in the decoder, where ${I_3^{R}}$ are the synthesised no-blur, normal-light and noise-free (Restoration) frames. We indicate ground truth frames of the corresponding tier by ${I_i^{GT}}$.

% Our VJT is designed to recover high-quality frames from low-quality input $T$ frames ${I^{LQ}}$, corrupted by varying degrees of video degradation, to achieve better restoration results. As shown in \cref{fig:arch}, the VJT can be divided into a shared Encoder and a Multi-tier Decoder. 

\subsection{Overall Framework}
In this work, we address the combined challenge of deblurring, enhancing low-light images, and denoising, framing it as a non-blind video restoration task. Utilizing synthetic datasets, we design a three-tiered decoder structure that progressively approaches different levels of ground truth. A feature fusion module links these tiers, facilitating the transition from shallow, singular feature learning to deep feature learning. To harmonize the three loss functions, we implement an adaptive weight scheme for more effective training.

% We represent the input frames, characterized by blurriness, low-light, and noise (denoted as Low Quality), as ${I^{LQ}}$, and the outputs of the decoder's three tiers as ${I_i^{R}}$, with ${I_3^{R}}$ representing the final restored frames (no blur, normal lighting, and noise-free, denoted as Restoration). The corresponding ground truth frames for each tier are indicated as ${I_i^{GT}}$.

Our Video Joint Task (VJT) framework aims to reconstruct high-quality frames from these low-quality inputs, ${I^{LQ}}$, which suffer from various degrees of video degradation. As illustrated in \cref{fig:arch}, the VJT architecture comprises a shared Encoder and a Multi-tier Decoder.

% \subsection{Encoder}
% Our encoder adopts a multi-blocks pyramid structure. Previous work has shown that the encoder structure with downsample parts is powerful at clustering features and reducing the number of parameters.

% Firstly, a spatial convolution is used for shallow feature extraction, a module utilised frequently in many previous images and video restoration works. We, therefore, receive $Z^{Shallow}$ as the output of the shallow feature.
% Then $Z^{Shallow}$ goes through a  spatio-temporal attention module. More details about attention module are in \cref{fig:attention} and the section on attention module.

% Next, we use deformable convolution\citep{dai2017deformable} to warp adjacent frames and use the optical flow obtained from the Spynet\citep{ranjan2017optical}.
% Finally, a downsample layer is passed. The above part is repeated three times to obtain the potential space $Z^{Latent}$.

\subsection{Encoder}

Our encoder features a multi-block pyramid structure, proven effective in feature clustering and parameter reduction in prior research.

Initially, spatial convolution extracts shallow features, yielding $Z^{Shallow}$. This is followed by a spatio-temporal attention module (detailed in \cref{fig:attention} and the Attention Module section). Next, we employ deformable convolution\citep{dai2017deformable} to align adjacent frames, leveraging optical flow data from the Spynet framework\citep{ranjan2017optical}. Finally, a downsample layer is passed. The above part is repeated three times to obtain the potential space $Z^{Latent}$.

% \subsection{Multi-tier Decoder}
% Our Multi-tier decoder consists of a three-tier network, with the number of tiers depending on the number of joint tasks. Each tier has three modules of different scales, each composed of three parts: upsample, attention and warping. After that each subsequent tier is followed by four attention modules and a reconstruction module which is a $3D$ convolution. The architecture of the Decoder is shown in \cref{fig:arch}.

% We take the output of the first tier $I_{DN}^R$ and calculate the $L_{DN}$ with the denoised target image $I^{GT}_{DN} $, which we hope will learn the denoised features in this tier. 

\subsection{Multi-tier Decoder}
Our Multi-tier Decoder is architected as a tri-tier neural network, tailored to the complexity of the joint tasks it addresses. Each tier within this network is segmented into three distinct modules, scaled variably to handle different aspects of the processing pipeline. These modules are intricately composed of three fundamental operations: upsampling to enhance resolution, attention mechanisms to focus on salient features, and warping to align and adjust frames temporally. Progressing beyond these initial operations, each tier sequentially integrates an additional four attention modules, further refining its focus and feature representation. This is followed by a reconstruction module, employing 3D convolutional layers for spatial-temporal data synthesis. The comprehensive design and structural intricacies of our Multi-tier Decoder are graphically delineated in \cref{fig:arch}, elucidating its layered and modular approach in tackling complex video restoration tasks.

The output of the first tier, $I_{DN}^R$, is compared against the denoised target image $I^{GT}_{DN}$ to compute the $L_{DN}$ loss,  aimed at enhancing the model's capability to extract and refine denoised features effectively.

%\subsubsection{Feature Fusion.} 
% We proposed a {\bf Feature Fusion} module between tiers. The first two tiers are used as an example. We wish to take the latent features from each module in the first tier and fuse them into the second tier. Firstly, We concatenate these latent features $Z_{i}^{1}$ with the output features $Z_{i-1}^{2}$ of the corresponding previous module in the second tier. Then we downsample them with a layer normalization and a linear layer. They are thus used as the respective input $\Tilde{Z}_{i}^{2}$, as shown in \cref{fig:fusion}. Similarly, there are also feature fusion modules between twelve attention module before the reconstruction.

% The output of the second tier $I_{LE\&DN}^R$ is used to calculate the $L_{LE\&DN}$ loss function with the ground truth image $I_{LE\&DN}^{GT}$ of the joint task denoising and Low-light enhancement.
% In the same way, its output features are downsampled and passed to the third tier for feature fusion.

% Finally, the output of the third tier $I_{DB\&LE\&DN}^R$ is the resulting video restoration that we ultimately want. We take it with the high-quality ground-truth video $I_{DB\&LE\&DN}^{GT}$ and calculate the$L_{DB\&LE\&DN}$ loss function.

We have innovatively introduced a \textbf{Feature Fusion} module to synergize the capabilities of consecutive tiers. Taking the first two tiers as a case study, our objective is to amalgamate the latent features extracted from each module of the first tier into the subsequent tier. This process involves a strategic concatenation of the latent features $Z_{i}^{1}$ from the first tier with the preceding module's output features $Z_{i-1}^{2}$ from the second tier. Subsequently, these combined features undergo downsampling, facilitated by layer normalization and linear transformation, to generate refined input features $\tilde{Z}_{i}^{2}$ for the next processing stage, as depicted in \cref{fig:fusion}. This feature fusion concept is also applied between the twelve attention modules preceding the reconstruction phase, enhancing the overall feature integration and processing flow.

\subsection{Attention Module}
\label{sec:attention}

% We compare several previous works on the Spatio-temporal attention mechanism and adopt the TMSA framework in VRT\citep{liang2022vrt}, \cref{fig:attention}, based on shifted window attention from SwinTransformer\citep{liu2021swin,liu2021video}.
% An attention module comprises $N$ parallel attentions and $M$ window multi-head self-attentions(W-MSA) from SwinTransformer\citep{liu2021swin}. The parallel attention is formed by window mutual multi-head attention (W-MMA) between adjacent frames and W-MSA. The structure of the attention module is shown in \cref{fig:attention}.
% $N$ depends on the number of video frames we input simultaneously. Since W-MMA is mutual attention between two frames, we need $N \geq T$ when inputting $T$ frames. $M$ modules with only W-MSA are to bring more attention back to the patches themselves, where $M$ is generally set to one-third to one-half of $N$.

In our exploration of advanced Spatio-temporal attention mechanisms, we conducted a comprehensive analysis of existing methodologies and ultimately integrated the TMSA (Temporal Multi-head Self-Attention) framework from VRT (referenced in Liang et al., 2022). This choice was influenced by the framework's innovative utilization of shifted window attention, a concept pioneered in the SwinTransformer (Liu et al., 2021).

The Attention module is architecturally composed of $N$ parallel attention mechanisms and $M$ window-based multi-head self-attentions (W-MSA). This dual-layered structure of attention is characterized by the integration of Window Mutual Multi-Head Attention (W-MMA), which operates between adjacent video frames, in conjunction with W-MSA. This unique arrangement enhances the module's capacity to simultaneously process temporal dynamics and spatial details, as detailed in \cref{fig:attention}.

The parameter $N$ is intrinsically tied to the number of video frames processed concurrently. Given that W-MMA facilitates mutual attention across pairs of frames, it becomes imperative that $N$ is at least equivalent to the number of frames $T$ being input, thereby ensuring comprehensive temporal coverage. Meanwhile, the $M$ modules, primarily focused on W-MSA, are designed to recalibrate and intensify attention towards individual frame patches. Typically, $M$ is calibrated to be between one-third and one-half of $N$, establishing a balanced attention distribution that prioritizes both inter-frame relationships and intra-frame details.

The Attention module uses the window mutual multi-head attention (WMMA) and window multi-head self-attention (WMSA) mechanisms to extract features. After that, it utilize MLP layers for dimensions reduction and feature fusion.
The formula for the Attention module is as follows, $Z_1$ is the input and $Z_2$ is the output:
\begin{equation}
    \begin{aligned}
    (X_1,X_2) &=LN(Z_1)\\
    Q_{1,2},K_{1,2},V_{1,2} &= MLP(X_{1,2})\\
    \widehat{Y}_1,\widehat{Y}_2 &= WMMA(Q_{2,1},K_{1,2},V_{1,2})\\
    Y_1,Y_2 &= WMSA(Q_{1,2},K_{1,2},V_{1,2})\\
    \widehat{Z}_1 &= MLP(Concatenate(\widehat{Y}_1,\widehat{Y}_2,Y_1,Y_2)) + Z_1\\
    Z_2 &= MLP(LN(\widehat{Z}_1)) + \widehat{Z}_1\\
    \end{aligned}
\end{equation}

% Figure of Data Synthesis pipeline
\begin{figure*}[htpb]
    \centering
    \includegraphics[width=0.95\linewidth]{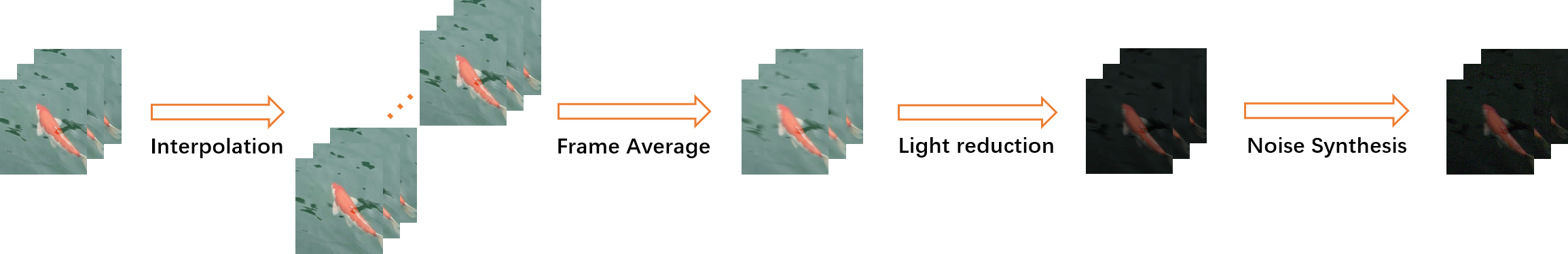}
    \caption{A concise diagram illustrates our data synthesis process. }
    \label{fig:data_gen}
\end{figure*}

\subsection{Loss Function}

% We use three loss functions corresponding to the three tiers of the framework: $L_{DN}$ for the single denoising tier, $L_{LE\&DN}$ for the joint low-light enhancement and denoising tier, and $L_{DB\&LE\&DN}$ for joint deblurring, illumination enhancement, and denoising tier. For consistency and uniformity of tasks, we use  Charbonnier loss\citep{charbonnier1994two} based on the L1 norm between the output $I^R_i(i \in \{DN, LE\&DN,DB\&LE\&DN\})$ of each tier and the corresponding target $I^{GT}_i$. Charbonnier loss is shown below:

In our multi-tier framework, we deploy triple set of loss functions, each tailored to a specific tier: $L_{DN}$ for denoising, $L_{LE\&DN}$ for the combined low-light enhancement and denoising, and $L_{DB\&LE\&DN}$ for the integrated deblurring, illumination enhancement, and denoising. To ensure task coherence and uniform performance metrics across these tiers, we utilize the Charbonnier loss, a robust variant of the L1 norm, to measure the discrepancy between the output $I^R_i(i \in \{DN, LE\&DN, DB\&LE\&DN\})$ of each tier and its corresponding ground truth $I^{GT}_i$. The Charbonnier loss function is defined as follows:

\begin{equation}
\label{loss}
    \begin{aligned}
        \mathcal{L}_{i}({I_i^R,I_i^{GT}}) = \sqrt{||I_i^R-I_i^{GT}||^2+\epsilon }  
    \end{aligned}
\end{equation}

where $\epsilon$ is a small positive constant that we set to $10^{-9}$.

% %\subsubsection{Adaptive Weight Scheme.} 
% Since our problem is a multi-task loss case, we are faced with the problem of how to adjust the weights of the loss functions. We, therefore, use an {\bf Adaptive Weighting Scheme} that refers to multi-task learning approaches\citep{kendall2018multi}.
% 
% % polish by chatgpt4
% % Considering the multi-task likelihood as a product of three individual task likelihoods, the likelihood is logarithmized according to maximum likelihood estimation, and the following approximation is obtained:
% By viewing the multi-task likelihood as a product of three individual task likelihoods, and applying logarithm transformation in line with maximum likelihood estimation, we derive the subsequent approximation:

%\subsubsection{Adaptive Weight Scheme.}
Given the multifaceted nature of our loss computation, encompassing diverse restoration tasks, we incorporate an \textbf{Adaptive Weighting Scheme} to dynamically balance the contribution of each individual loss function. This scheme draws inspiration from advanced multi-task learning strategies \citep{kendall2018multi} and is formulated by considering the multi-task likelihood as a composite of the individual task likelihoods. By applying logarithmic transformation aligned with the principles of maximum likelihood estimation, we obtain the following adaptive loss formulation:

In addressing the complexities of multi-task loss in our model, the challenge lies in optimally adjusting the weights of individual loss functions. To resolve this, we implement an \textbf{Adaptive Weighting Scheme}, grounded in established multi-task learning methodologies \citep{kendall2018multi}. This approach conceptualizes the multi-task likelihood as the product of the likelihoods for each individual task. Through the application of a logarithmic transformation, consistent with maximum likelihood estimation principles, we derive an effective approximation:

\begin{equation}
\label{multiloss}
    \begin{aligned}
     \mathcal{\widehat{L}}_{i} =\frac{1}{2 \sigma_{i}^{2}} \mathcal{L}_{i}({I_i^R,I_i^{GT}})+\log \sigma_{i}
    \end{aligned}
\end{equation}
where $\sigma_i$  denotes the model's observation noise parameter of each tier. The Adam optimizer updates it to get better weights.
Then, based on Liebel's method \citep{liebel2018auxiliary}, we modify the 
logarithmic form so that it would be stably convergent. Hence, we use:
\begin{equation}
\label{multiloss_2}
    \begin{aligned}
     \mathcal{\widetilde{L}}_{i} =\frac{1}{2 \sigma_{i}^{2}} \mathcal{L}_{i}({I_i^R,I_i^{GT}})+\log (1+\sigma_{i}^{2}).
    \end{aligned}
\end{equation}
The overall loss function is:
\begin{equation}
\label{multiloss_3}
    \begin{aligned}
     \mathcal{L} =\mathcal{\widetilde{L}}_{DN} +\mathcal{\widetilde{L}}_{LE\&DN} +\mathcal{\widetilde{L}}_{DB\&LE\&DN}
    \end{aligned}.
\end{equation}
It can adjust the weights faster to achieve better training results.

\section{Our MLBN Dataset}
\label{sec:data}

A real lowlight-blur-noise dataset is difficult to collect because we cannot make the same random camera movements in the same scene, in normal light and low light.  If we specify the camera motion pattern with a specific trace, the blur kernel is also determined, which does not work for the dataset. So we choose to generate our dataset.
After systematic research, the existing low-light datasets, either under a fixed lens such as SMID dataset \citep{chen2019seeing}, or under a uniform horizontal movement such as SDSD dataset \citep{wang2021seeing}, were too simple in their motion to be suitable for our lowlight-blur-noise requirements. Our dataset is therefore intended to be generated based on the deblur dataset.

Our Multi-scenes Lowlight-Blur-Noise Dataset (MLBN Dataset)  contains three main types of scenes:  indoor scenes, nighttime outdoor scenes,  and daytime outdoor scenes. The RealBlur dataset \citep{rim_2020_realblur} is chosen as the basis for the indoor scenes and some of the night scenes. Besides, the day scenes and the other night scenes are generated from our own filming and 4K videos on YouTube. A total of 165 scenes containing 25 longer daytime outdoor videos are used for training, and another 30 scenes are used for testing.

%\subsection{Data Synthesis Progress}

\subsection{Adding Motion Blur.}
Firstly, in order to circumvent the relatively sharp blur caused by the averaging of a small number of frames, we initially employ the frame interpolation method RIFE to augment the original 60FPS video to 1920FPS, thereby ensuring the continuity of the blur.
Then for 4k HD videos $I_{4k}^{GT}$, we resize them to $1280\times720$ $I_{720}^{GT}$. This step could effectively reduce the noise in videos. We centrally crop them to the same size as the RealBlur dataset for easy training and testing. 
After that, we perform the averaging process, ensuring the continuity of the blurriness.
The most common method of generating blurry images is from the following blur model:
\begin{equation}
\label{blur1}
    \begin{aligned}
     I^{B} = KI^{GT}+n
    \end{aligned}
\end{equation}
$K$ is a large sparse matrix containing a local blur kernel at each row, and $n$ is noise.
However, since the inversion of $K$ is ill-posed and not universal, we choose to generate the blurry image $I^B$ based on method \citep{SeungjunNah2016DeepMC}, using multi-frame temporal information:
\begin{equation}
\label{blur2}
    \begin{aligned}
     I^{B} = g(\frac{1}{T} \sum_{t=1}^{T}g^{-1}(I_t^{GT}))
    \end{aligned}
\end{equation}
where $g$ is the camera response function (CRF).
Next, we take 10 seconds (total 19200 frames) of video from each scene, make a blurred image from 160 consecutive frames, and the middle frame is taken as the ground truth image so that these scenes will have 120 paired frames $I^{B} \& I^{GT}$.

\subsection{Reducing Illumination.}
Starting with $I^{Blur}$, to be able to reduce illumination, we choose the Retinex model. Retinex theory is based on the central assumption that the image can be decomposed into the reflectance and the illumination components:

\begin{equation}
\label{retinex}
    \begin{aligned}
     I^{B} = R \circ H
    \end{aligned}
\end{equation}
where $I$ is the original image, $R$ is the reflected component and $H$ is the incident component, i.e. the illumination. Generally, $H$ is enhanced to obtain a normally illuminated image.

We implement the inverse algorithm based on the Retinex method LIME\citep{guo2016lime}. After decomposing the normally illuminated image into $R$ and $H$, $H$ is weakened by gamma correction to obtain the new low-illumination map $\hat{H}$. In addition, unlike LIME, we also perform a slight attenuation of the $R$ component to obtain $\hat{R}$, which is shown to produce an image that better matches the real scene in experiments. We obtain the low-light blur image as: 
\begin{equation}
\label{gamma correction}
    \begin{aligned}
    \hat{H} = H^{\gamma_1}, \hat{R} = R^{\gamma_2}
    \end{aligned}
\end{equation}
and
\begin{equation}
\label{retinex_inverse}
    \begin{aligned}
    I^{L\&B} = \hat{R} \circ \hat{H}.
    \end{aligned}
\end{equation}

We also use the augmented Lagrangian multiplier (ALM) algorithm and possible weighting strategies to accelerate it.

\subsection{Synthesise Noise.}
Synthetic datasets are typically generated using the widely assumed additive white Gaussian noise (AWGN). However, they perform poorly when applied to real camera videos. This is mainly because AWGN is insufficient to model real camera noise, which is signal-dependent. Therefore, we adapt the CycleISP model that simulates a large number of transformations in a camera imaging pipeline for noise generation.

% Visual Picture1
\begin{figure*}[htb]
    \setlength{\abovecaptionskip}{1.5pt}
    \centering
    
    % 第一行
    \begin{subfigure}{.25\textwidth}
        \centering
        \includegraphics[width=0.95\linewidth]{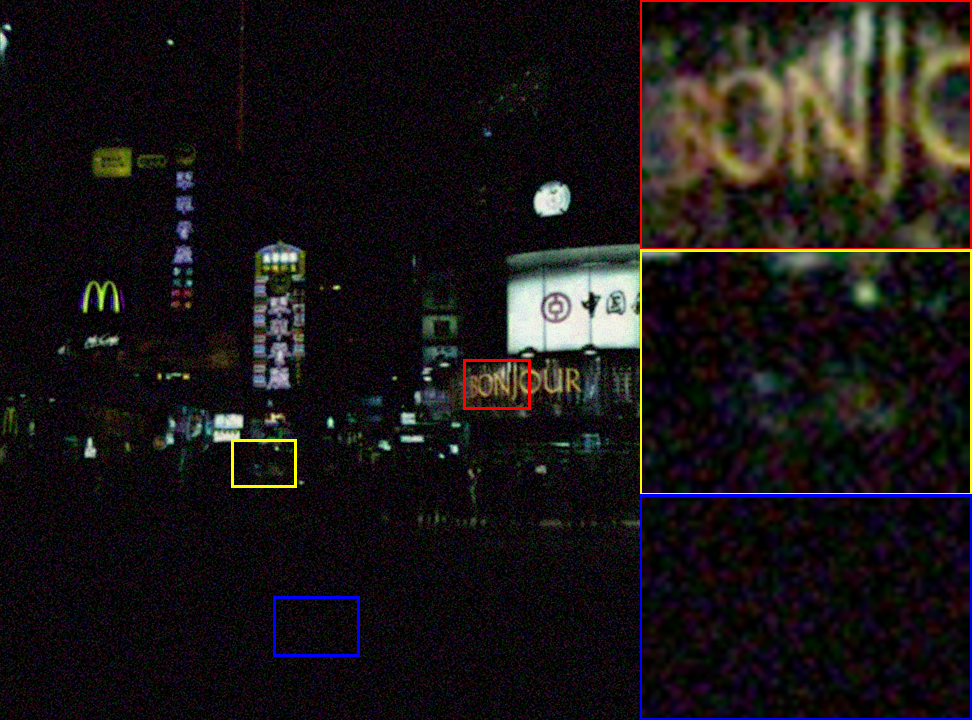}
        \caption{Lowlight-Blur-Noise Images}
        %\label{fig:lq1}
    \end{subfigure}%
    \begin{subfigure}{.25\textwidth}
        \centering
        \includegraphics[width=0.95\linewidth]{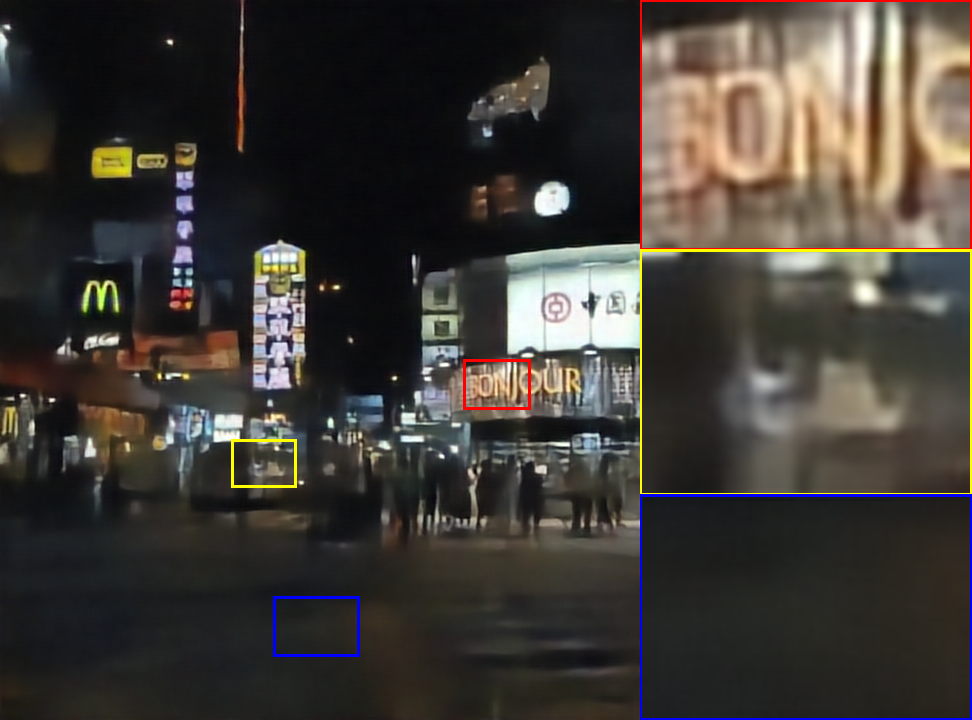}
        \caption{ESTRNN \citeauthor{zhong2020estrnn}}
        %\label{fig:ESTRNN1}
    \end{subfigure}%
    \begin{subfigure}{.25\textwidth}
        \centering
        \includegraphics[width=0.95\linewidth]{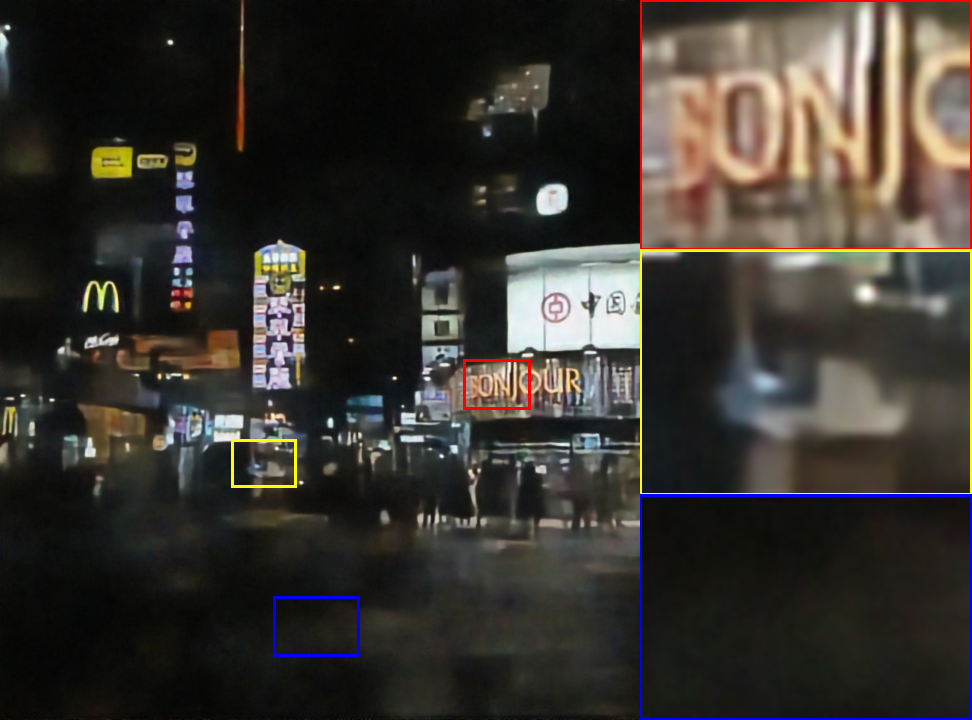}
        \caption{FGST \citeauthor{fgst}}
        %\label{fig:fgst1}
    \end{subfigure}%
    \begin{subfigure}{.25\textwidth}
        \centering
        \includegraphics[width=0.95\linewidth]{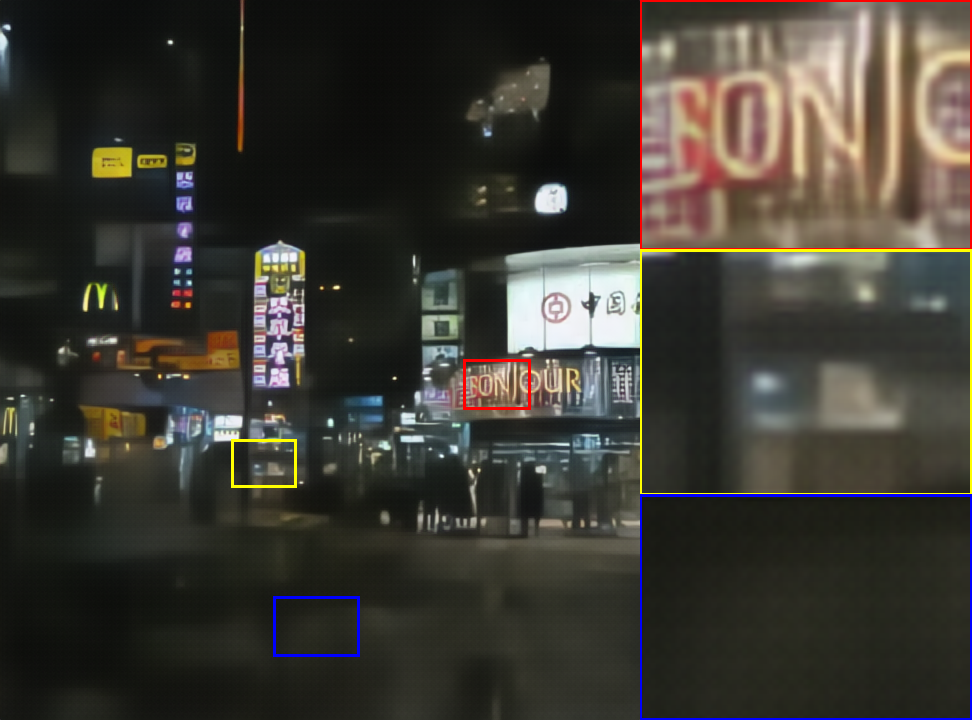}
        \caption{LEDNet \citeauthor{zhou2022lednet}}
        %\label{fig:lednet1}
    \end{subfigure}%
    
    % 第二行
    \begin{subfigure}{.25\textwidth}
        \centering
        \includegraphics[width=0.95\linewidth]{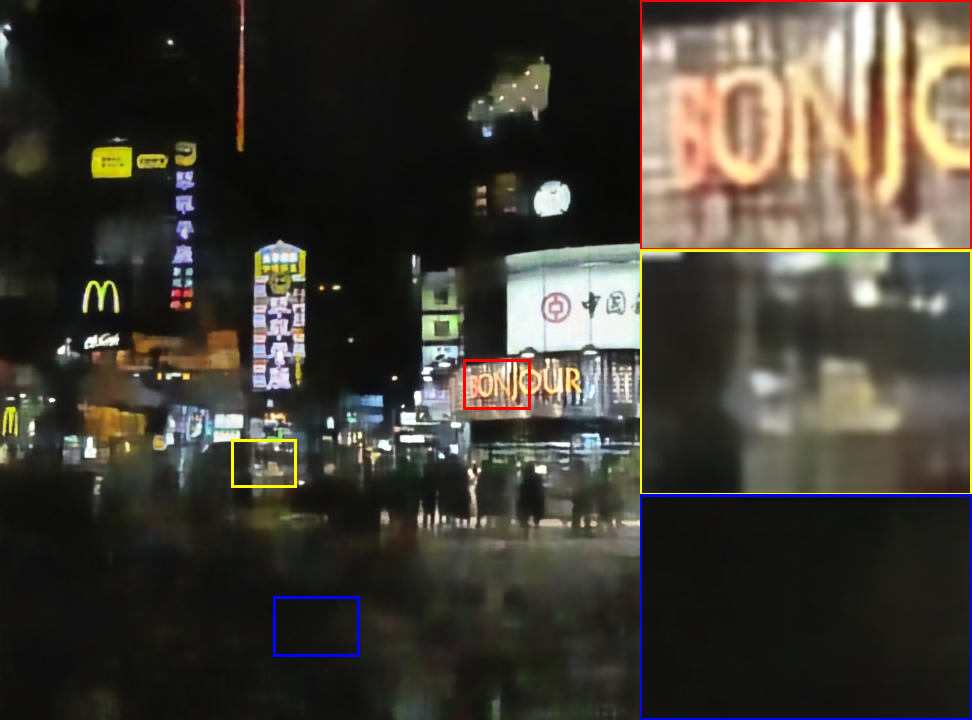}
        \caption{VRT \citeauthor{liang2022vrt}}
        %\label{fig:vrt1}
    \end{subfigure}%
    \begin{subfigure}{.25\textwidth}
        \centering
        \includegraphics[width=0.95\linewidth]{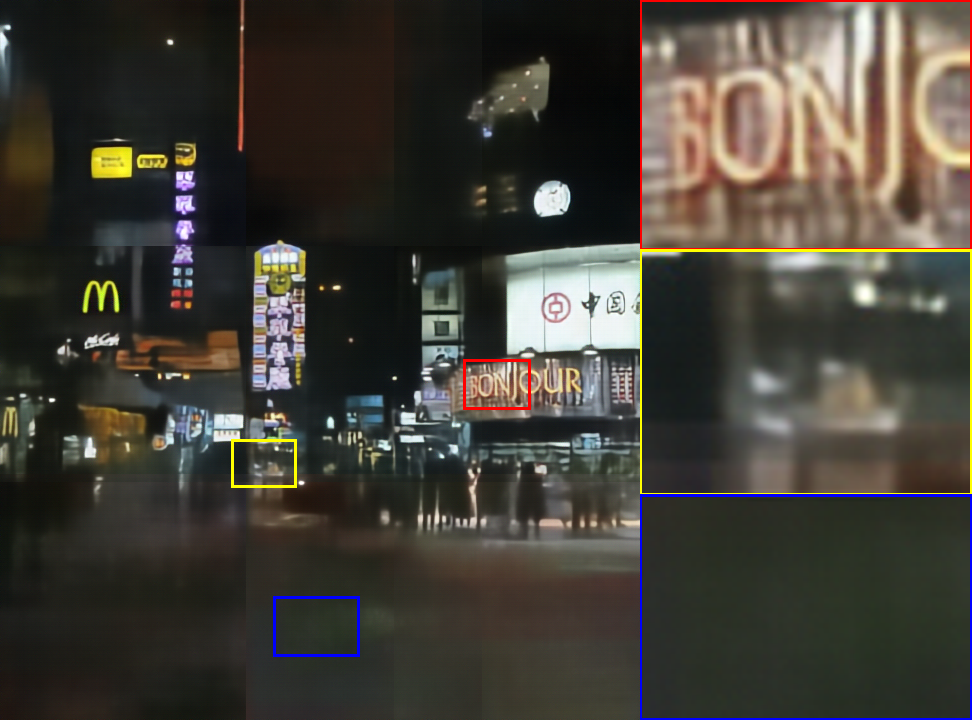}
        \caption{RVRT \citeauthor{liang2022recurrent}}
        %\label{fig:rvrt1}
    \end{subfigure}%
    \begin{subfigure}{.25\textwidth}
        \centering
        \includegraphics[width=0.95\linewidth]{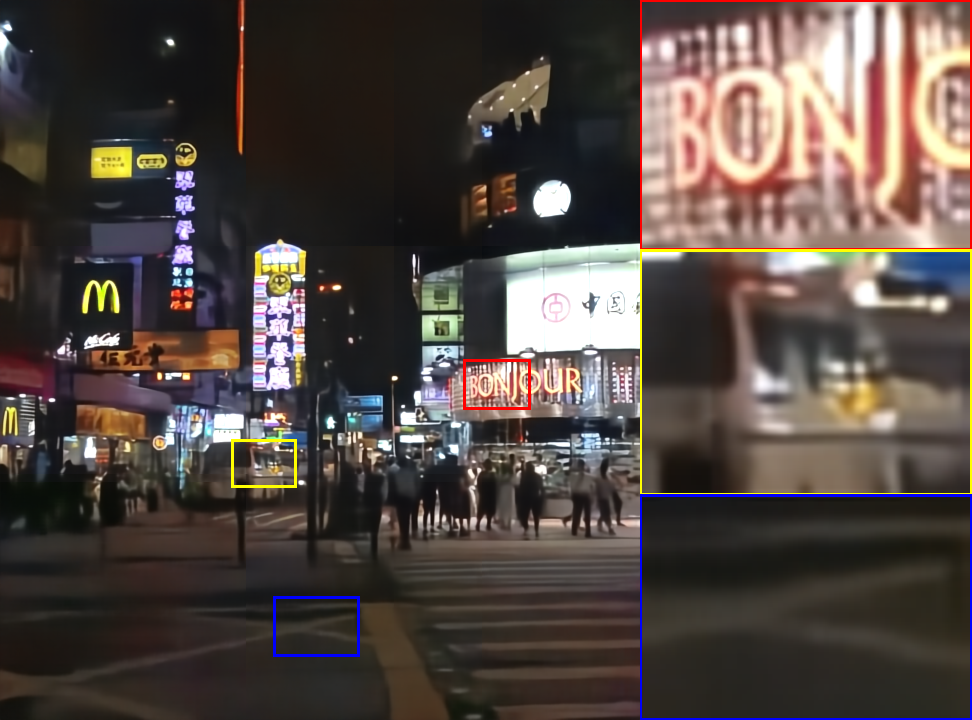}
        \caption{Ours Method}
        %\label{fig:ours1}
    \end{subfigure}%
    \begin{subfigure}{.25\textwidth}
        \centering
        \includegraphics[width=0.95\linewidth]{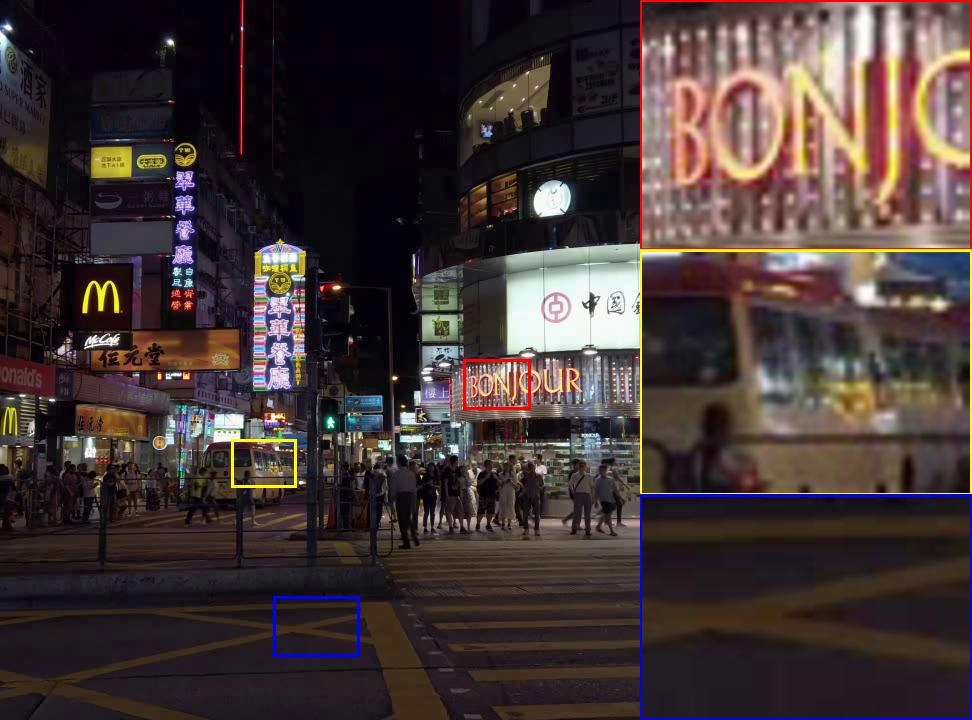}
        \caption{Ground Truth}
        %\label{fig:gt1}
    \end{subfigure}
    
    \caption{Qualitative visual comparisons on a frame of a night-time outdoor street scene in our MLBN dataset. Our approach demonstrates enhanced capabilities in restoring nighttime illumination, deblurring, and denoising. We mask it to show it more visually.}
    \label{fig:result1}
\end{figure*}

\section{Experiments}
\subsection{Experiments Settings}

We train all models with the learning rate initialized as $4e-4$ and adapted the cosine annealing schedule \citep{loshchilov2016sgdr} with a minimum learning rate of $1e-7$. We use the adam optimizer \citep{kingma2014adam} for training with $\beta_1=0.9, \beta_2=0.99$ for a total of 100k iterations. In the adaptive weight scheme, we initialize all ${\sigma}_{i}$ to 1.
%, resulting in an initial coefficient of $0.5$, while the bias term in the logarithm $log(1+ \sigma_i^2)$ corresponds to 0.
In addition, for the network, we use a pyramid structure with $3$ downsample and upsample layers, followed by $4$ equally sized attention modules. 
%We set the latent channel to uniform $48$ in the pyramid structure and $60$ in the following attention modules.

Each input sequence has $6$ frames, and the patch size is $192 \times 192$, which is cropped in pairs randomly. Patches are also transformed by rotations of $90^{\circ}, 180^{\circ}, 270^{\circ} $ and horizontal flipping at random. And the window size of each patch is $6 \times 8\times 8$. 

In our model, we use a total of $7$ layers of Attention-Warping modules of different scales and $4$ layers of Attention modules of the same size. The latent channel numbers are $48, 60$ respectively.
For each of the $7$ Attention-Warping modules, $N=6$ layers with the window mutual multi-head attention (WMMA) and window multi-head self-attention (WMSA) mechanism and $M=2$ layers with only WMSA mechanism were used. Each of the $4$ Attention modules contains $4$ layers of WMSA mechanism module, for a total of $16$layers. The head numbers of multi-head attentions are all $6$ and the deformable groups are $16$. We also use DropPath in the model for better training.

We trained our model with batch size $4$ on a server with $2$ Intel Xeon Gold $6238$R CPUs and $4$ NVIDIA RTX A$6000$ GPUs.
We adopt the PSNR and SSIM metrics for quantitative evaluation.

% Visual Picture2
\begin{figure*}[htb]
    \setlength{\abovecaptionskip}{1.5pt}
    \centering
    
    % 第一行
    \begin{subfigure}{.25\textwidth}
        \centering
        \includegraphics[width=0.95\linewidth]{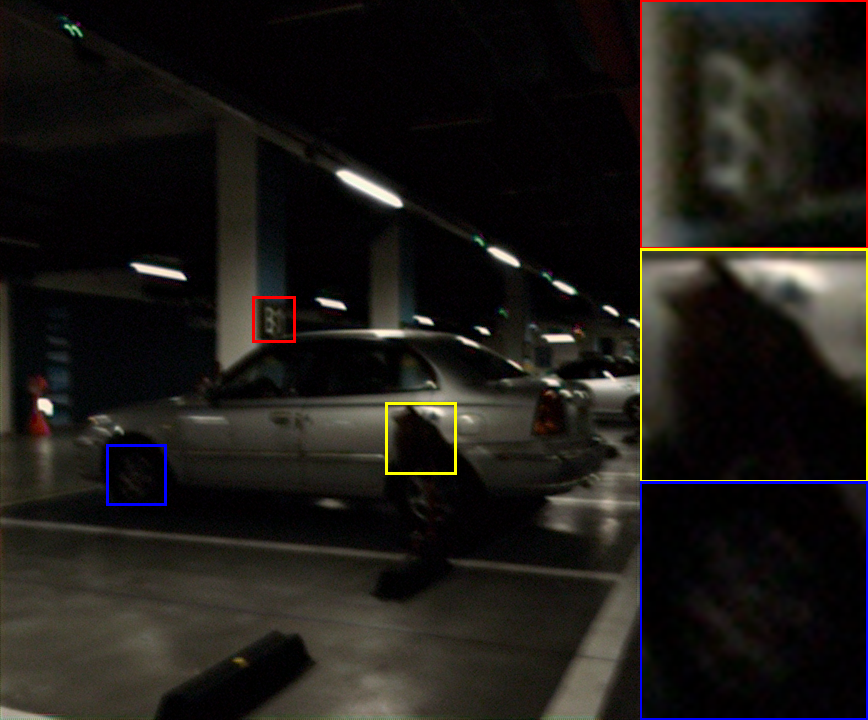}
        \caption{Lowlight-Blur-Noise Images}
        %\label{fig:lq1}
    \end{subfigure}%
    \begin{subfigure}{.25\textwidth}
        \centering
        \includegraphics[width=0.95\linewidth]{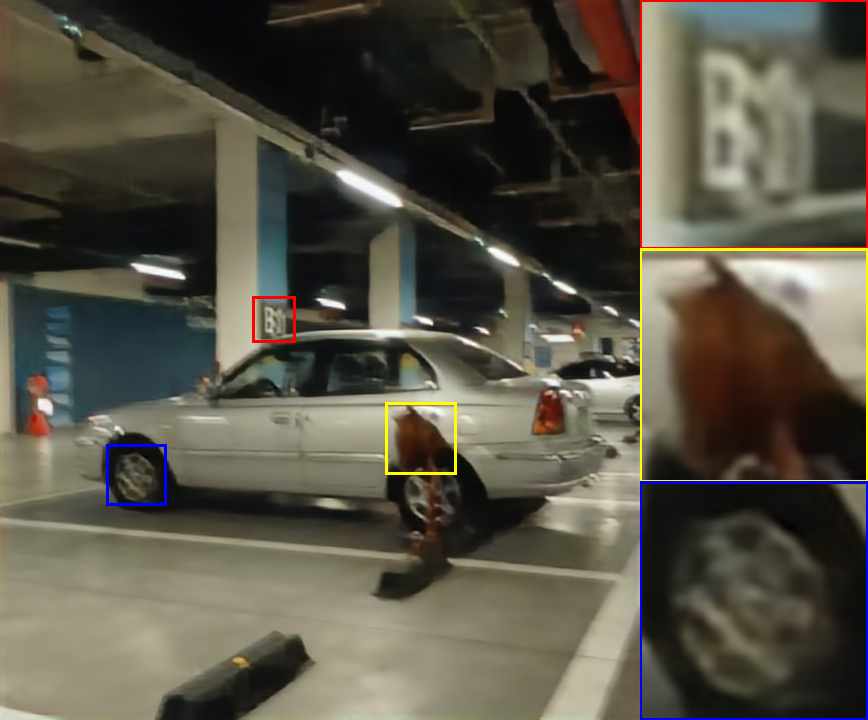}
        \caption{ESTRNN \citeauthor{zhong2020estrnn}}
        %\label{fig:ESTRNN1}
    \end{subfigure}%
    \begin{subfigure}{.25\textwidth}
        \centering
        \includegraphics[width=0.95\linewidth]{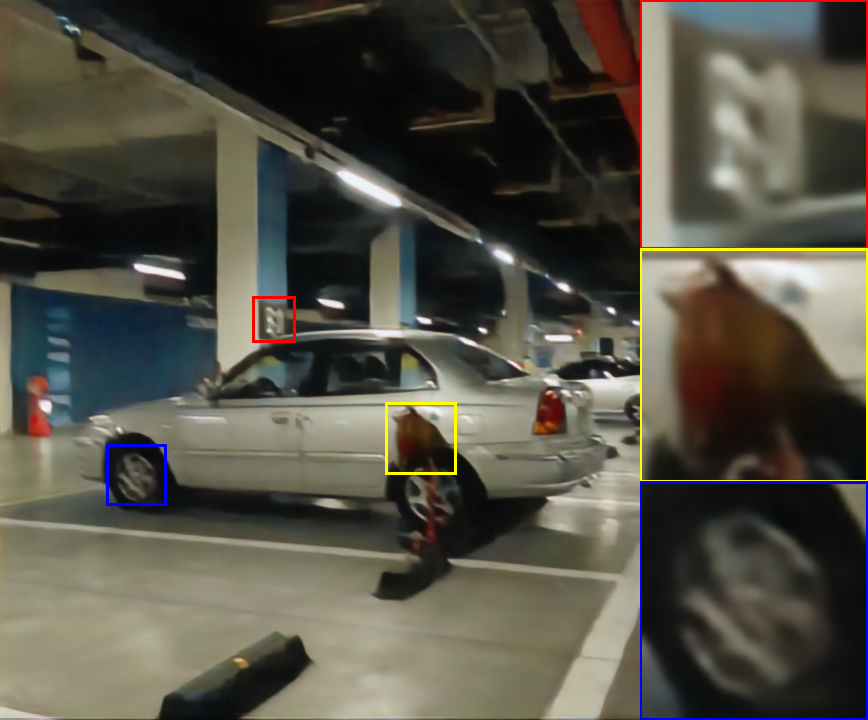}
        \caption{FGST \citeauthor{fgst}}
        %\label{fig:fgst1}
    \end{subfigure}%
    \begin{subfigure}{.25\textwidth}
        \centering
        \includegraphics[width=0.95\linewidth]{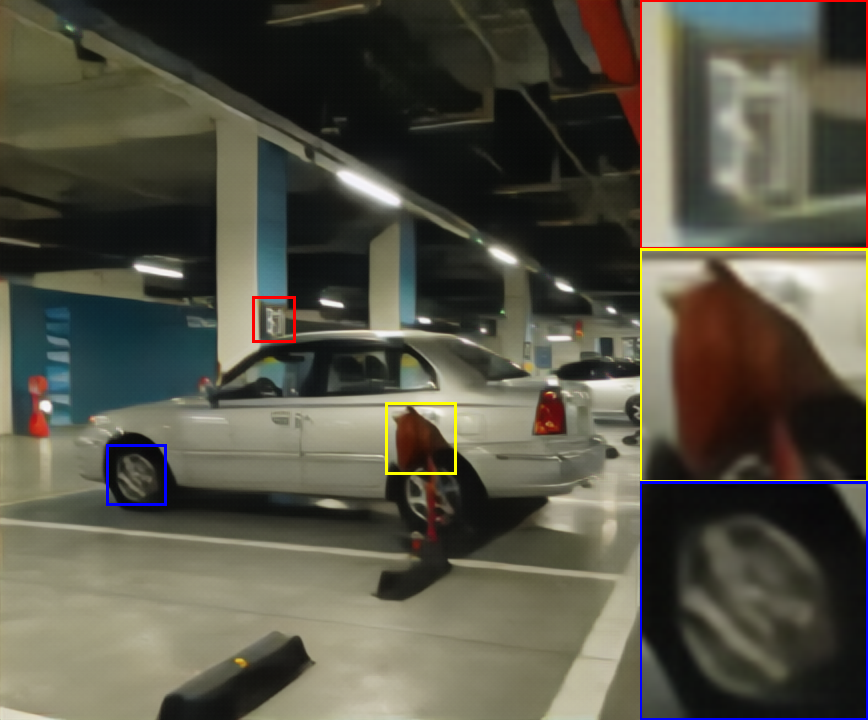}
        \caption{LEDNet \citeauthor{zhou2022lednet}}
        %\label{fig:lednet1}
    \end{subfigure}%
    
    % 第二行
    \begin{subfigure}{.25\textwidth}
        \centering
        \includegraphics[width=0.95\linewidth]{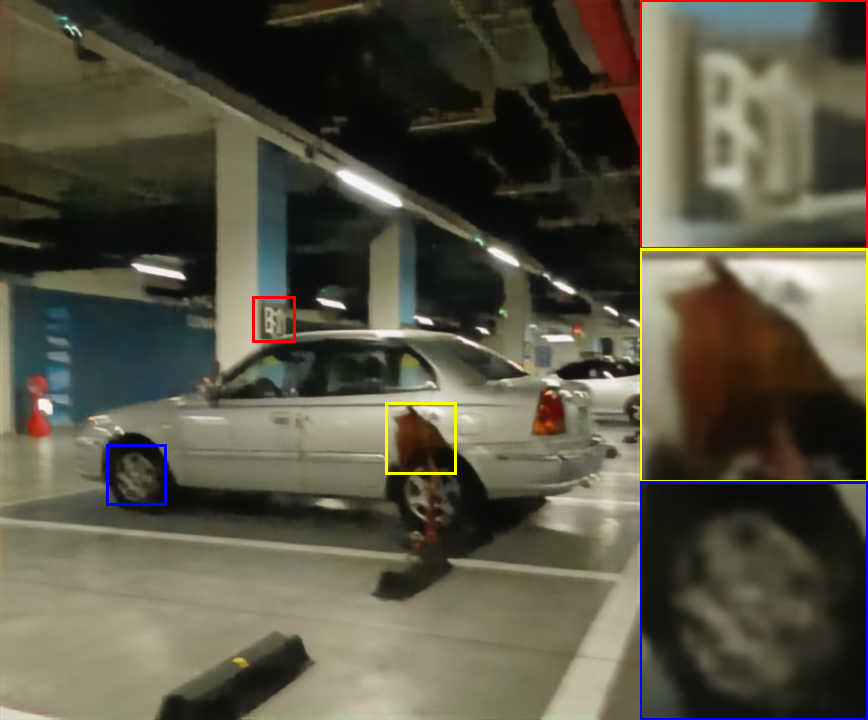}
        \caption{VRT \citeauthor{liang2022vrt}}
        %\label{fig:vrt1}
    \end{subfigure}%
    \begin{subfigure}{.25\textwidth}
        \centering
        \includegraphics[width=0.95\linewidth]{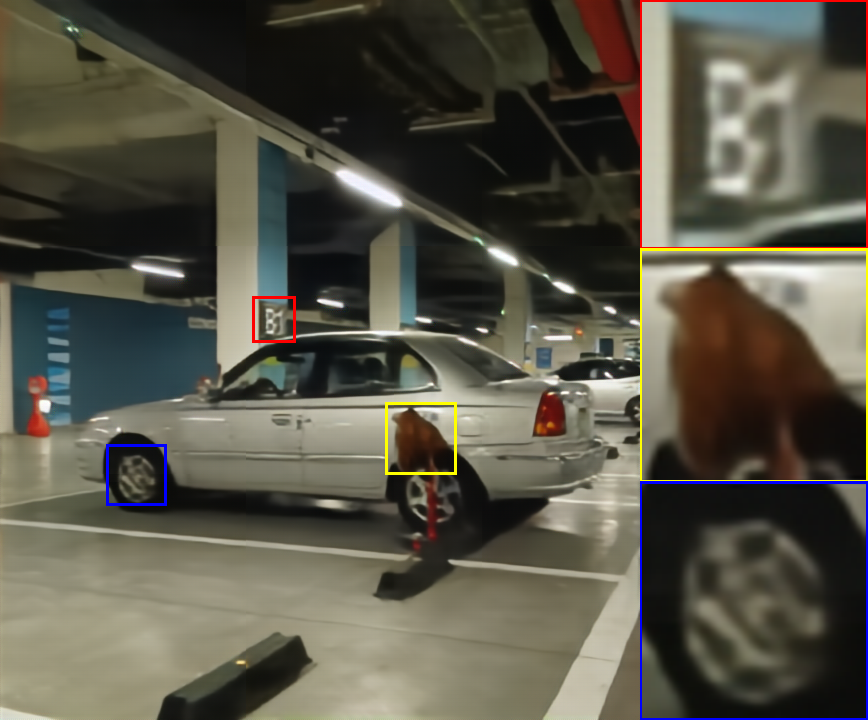}
        \caption{RVRT \citeauthor{liang2022recurrent}}
        %\label{fig:rvrt1}
    \end{subfigure}%
    \begin{subfigure}{.25\textwidth}
        \centering
        \includegraphics[width=0.95\linewidth]{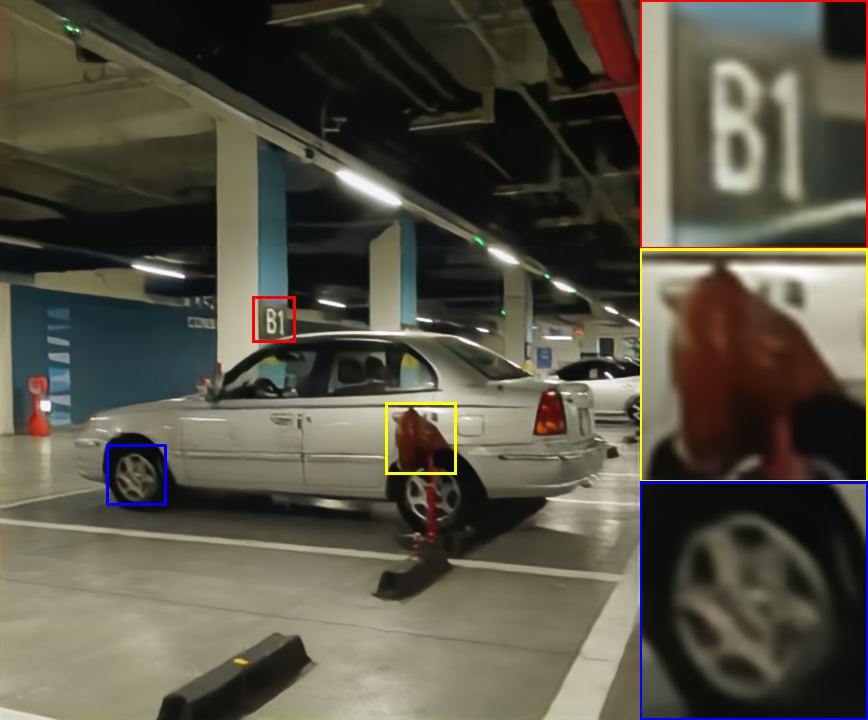}
        \caption{Ours Method}
        %\label{fig:ours1}
    \end{subfigure}%
    \begin{subfigure}{.25\textwidth}
        \centering
        \includegraphics[width=0.95\linewidth]{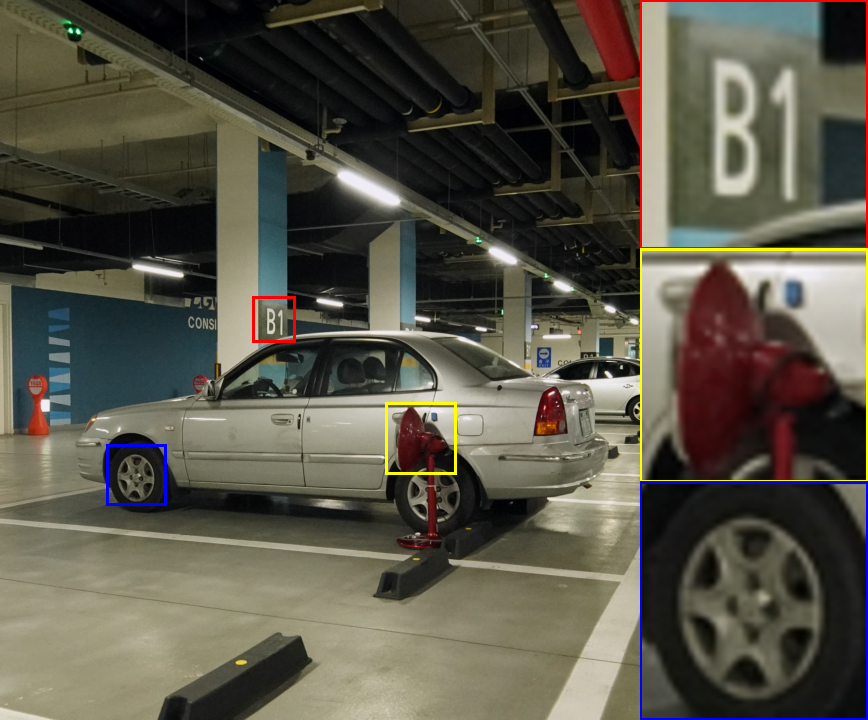}
        \caption{Ground Truth}
        %\label{fig:gt1}
    \end{subfigure}
    
    \caption{Qualitative visual comparisons on a frame of a indoor scene in our MLBN dataset. Our method offers improved {\bf Deblurring and Denoising} effects, such as the {\bf 'B1' sign on the pillar}, as well as enhanced illumination effects, like the red reflective mirror.We zoom in to show it more visually.}
    \label{fig:result2}
\end{figure*}

% Visual Picture3
\begin{figure*}[htb]
    \setlength{\abovecaptionskip}{1.5pt}
    \centering
    
    % 第一行
    \begin{subfigure}{.25\textwidth}
        \centering
        \includegraphics[width=0.95\linewidth]{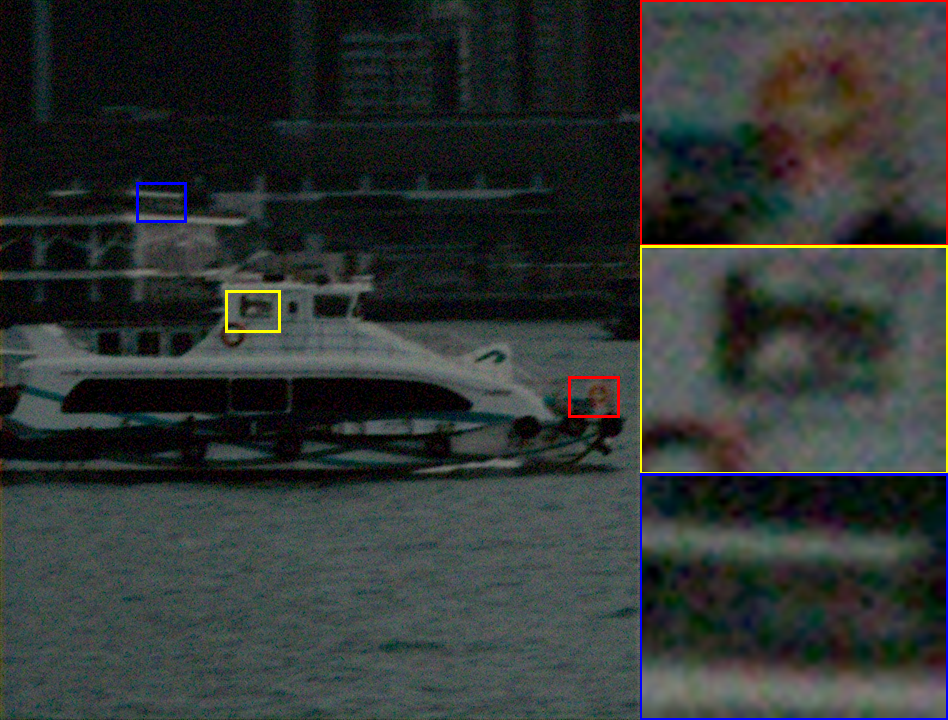}
        \caption{Lowlight-Blur-Noise Images}
        %\label{fig:lq1}
    \end{subfigure}%
    \begin{subfigure}{.25\textwidth}
        \centering
        \includegraphics[width=0.95\linewidth]{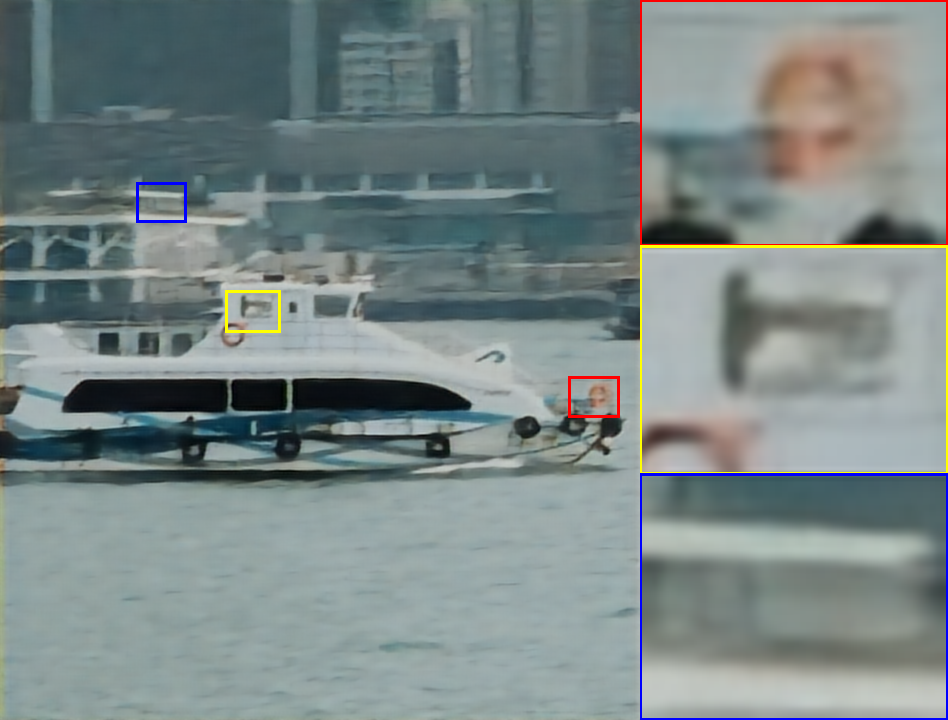}
        \caption{ESTRNN \citeauthor{zhong2020estrnn}}
        %\label{fig:ESTRNN1}
    \end{subfigure}%
    \begin{subfigure}{.25\textwidth}
        \centering
        \includegraphics[width=0.95\linewidth]{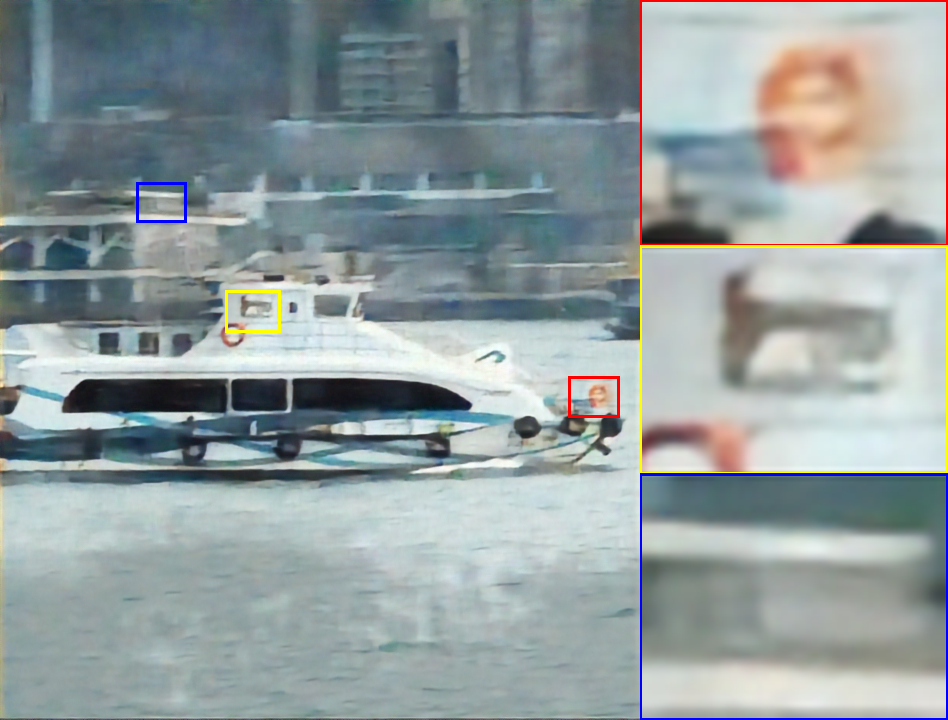}
        \caption{FGST \citeauthor{fgst}}
        %\label{fig:fgst1}
    \end{subfigure}%
    \begin{subfigure}{.25\textwidth}
        \centering
        \includegraphics[width=0.95\linewidth]{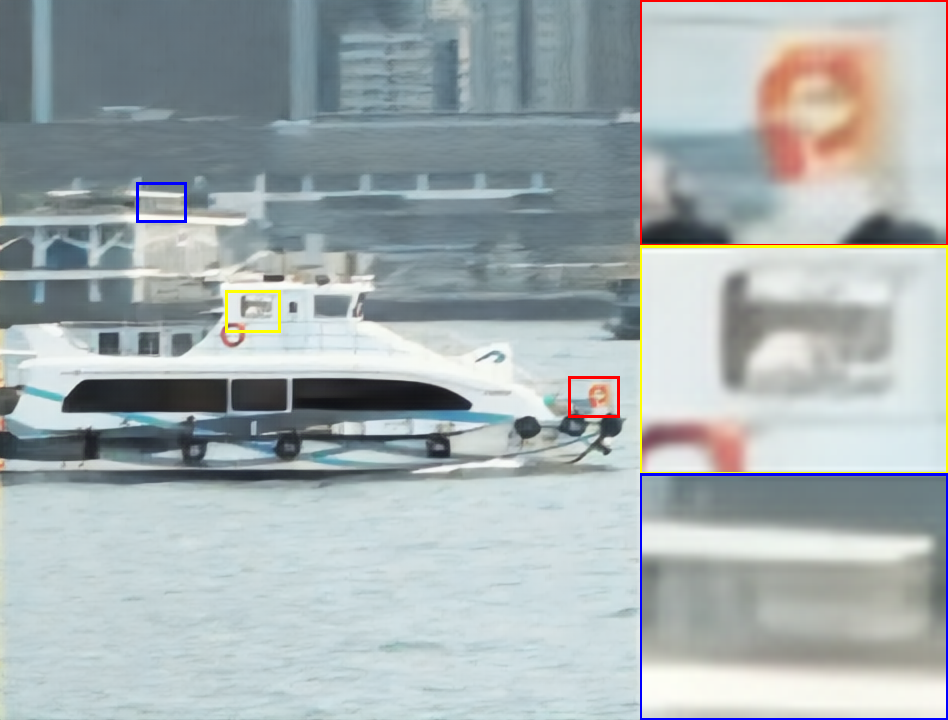}
        \caption{LEDNet\citeauthor{zhou2022lednet}}
        %\label{fig:lednet1}
    \end{subfigure}%
    
    % 第二行
    \begin{subfigure}{.25\textwidth}
        \centering
        \includegraphics[width=0.95\linewidth]{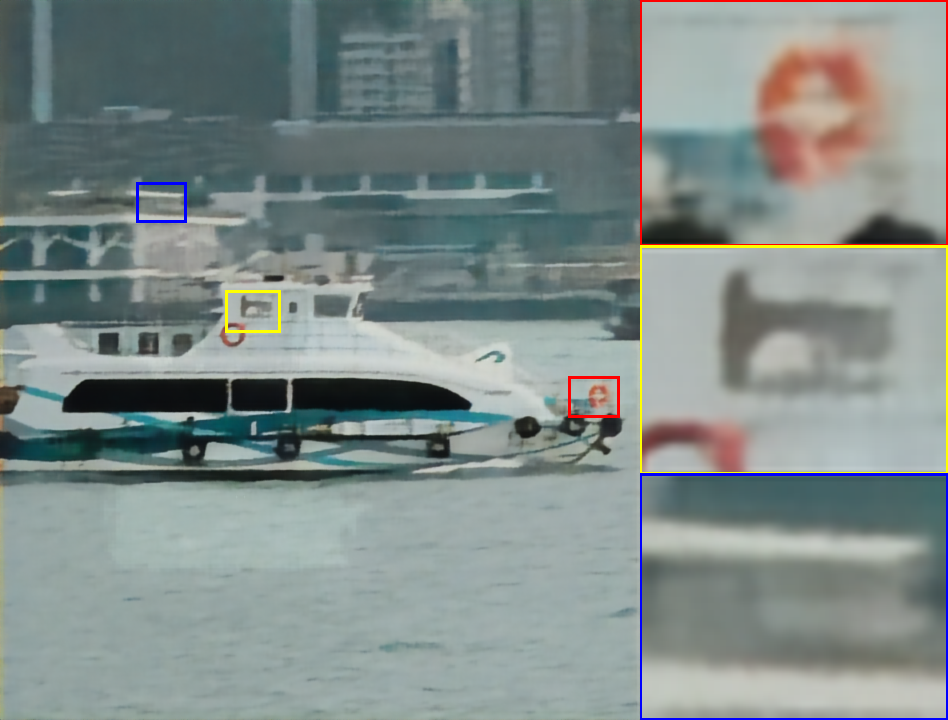}
        \caption{VRT\citeauthor{liang2022vrt}}
        %\label{fig:vrt1}
    \end{subfigure}%
    \begin{subfigure}{.25\textwidth}
        \centering
        \includegraphics[width=0.95\linewidth]{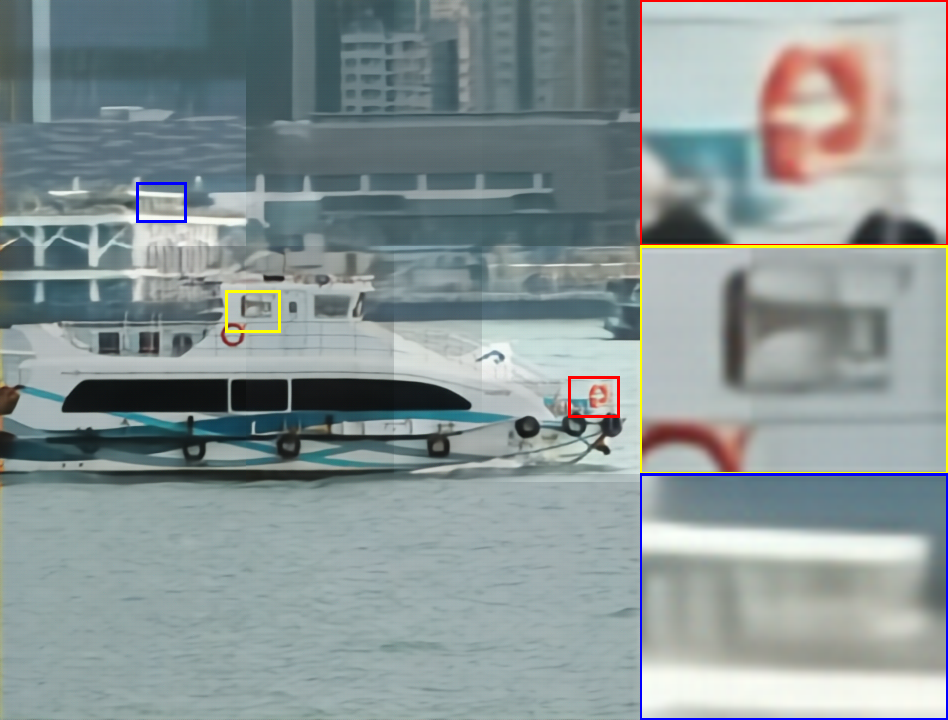}
        \caption{RVRT \citeauthor{liang2022recurrent}}
        %\label{fig:rvrt1}
    \end{subfigure}%
    \begin{subfigure}{.25\textwidth}
        \centering
        \includegraphics[width=0.95\linewidth]{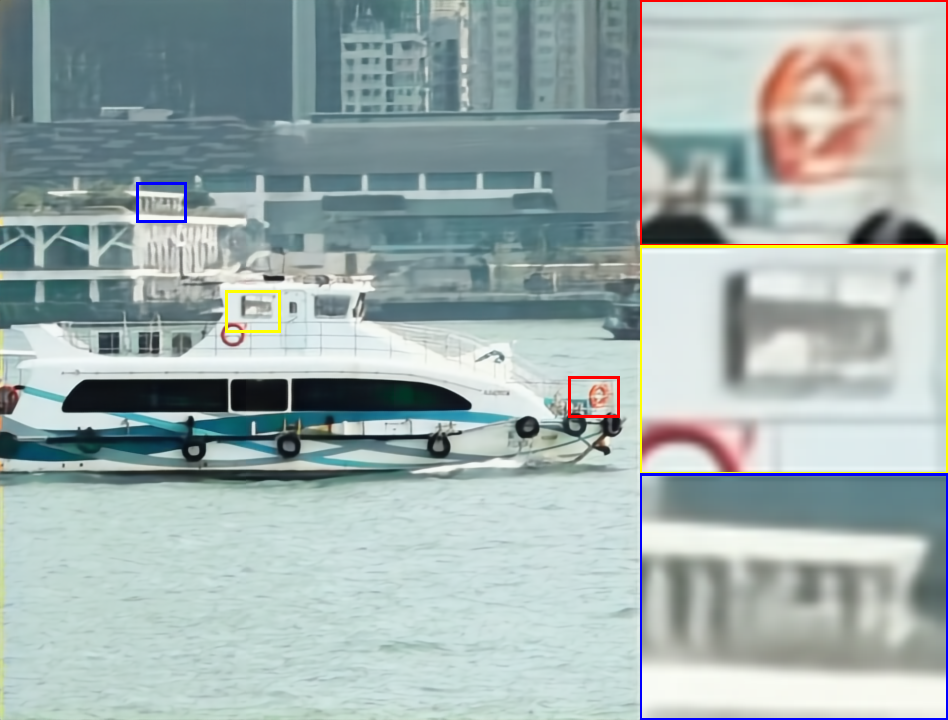}
        \caption{Ours Method}
        %\label{fig:ours1}
    \end{subfigure}%
    \begin{subfigure}{.25\textwidth}
        \centering
        \includegraphics[width=0.95\linewidth]{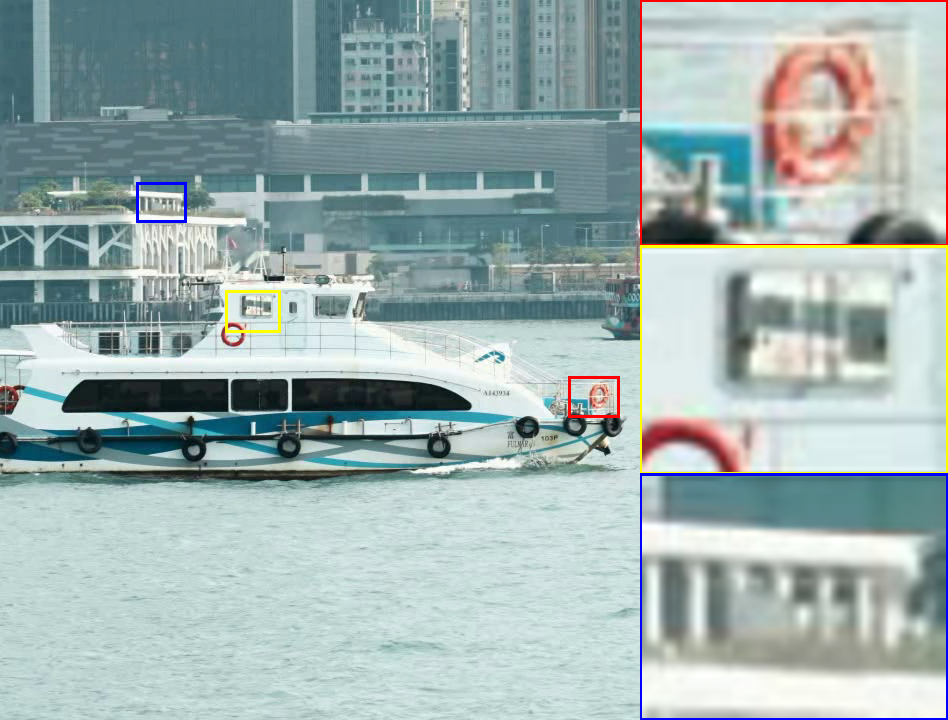}
        \caption{Ground Truth}
        %\label{fig:gt1}
    \end{subfigure}
    
    \caption{Qualitative visual comparisons on a frame of a daytime outdoor scene in our MLBN dataset. Our method delivers {\bf Enhanced Illumination and Denoising } results, exemplified by the {\bf lifebuoy on the bow of the boat}. We zoom in to show it more visually.}
    \label{fig:result3}
\end{figure*}

\subsection{Evalution on Our MLBN Dataset}

\subsubsection{Baselines.}
As there is no similar end-to-end method for the joint tasks of video deblurring, low-light enhancement and denoising before, we choose several state-of-the-art end-to-end models for video deblurring as our baselines. 
ESTRNN\citep{zhong2020estrnn} designed a global spatio-temporal attention module and used RDB-based RNN cells. 
FGST\citep{fgst} proposed an optical flow-guided attention mechanism FGSW-MSA to expand the time-domain receptive field.
LEDNet\citep{zhou2022lednet} designed a LE-Encoder for light enhancement and a Deblurring Decoder.
VRT\citep{liang2022vrt} proposed neighbour frame attention and parallel warping with optical flow. The SOTA performance video transformer for deblurring is RVRT \citep{liang2022recurrent}
, which propese a recurrent framework based on VRT.
With their released code, these models are well-retrained on our Multi-scene-Lowlight-blur-noise (MLBN) dataset. 

The comparative methods were trained until convergence, utilizing only the ground truth data from the final stage, due to their inherent structural configurations.

% PSNR & SSIM Table
\begin{table}[htpb]
\setlength{\abovecaptionskip}{0.1cm}
\setlength{\belowcaptionskip}{-0.3cm}
	\caption{ Quantitative results with PSNR, SSIM metrics on our MLBN dataset. Videos with higher scores have better visual quality. The \textcolor{red}{BEST} results are in red, whereas the \textcolor{blue}{SECOND BEST} ones are in blue. Our method achieves the best results on the MLBN dataset.}
	\centering
	{\begin{tabular}{c|c c }
			\hline		
			Methods    & PSNR$\uparrow$  & SSIM $\uparrow$     \\ \hline
		ESTRNN \citep{zhong2020estrnn}   & {23.72} &  {0.7600}   \\
            FGST \citep{fgst}  & {22.9434}  & {0.6997}  \\
			LEDNet \citep{zhou2022lednet} & {23.98}   &  {0.7234}  \\
		  VRT \citep{liang2022vrt}   & {23.37} & {0.7430}  \\
            RVRT \citep{liang2022recurrent}   & \textcolor{blue}{24.71} & \textcolor{blue}{0.7612}  \\
		Ours       & \textcolor{red}{25.45} & \textcolor{red}{0.8083}   \\ \hline
	\end{tabular}}
	\label{tab:psnr} 
\end{table}

\subsubsection{Quantitative Analysis.} 
\Cref{tab:psnr} shows the quantitative comparison results on the MLBN dataset. 
VJT achieves the best performance, outscoring the second-best method RVRT in terms of PSNR and SSIM by  $0.74dB$ and $0.0471$. Compared to the joint image deblurring and illumination enhancement method LEDNet, we have even a $1.47dB$ improvement in PSNR. The total number of parameters of VJT is $12.17$M, and the average running time of a frame is $692$ ms.
Because we have set the channel size(48,60) of our intermediate layer to half of VRT(96,120), although our framework appears large from the graph, in reality, both our encoder and each tier of decoder are less than half of VRT, and our total number of parameters is also smaller both than VRT and RVRT.
More information about the model size and runtime of compared SOTA models can be found in \Cref{tab:runtime}.

% Runtime & Model size Table
\begin{table}[htpb]
\setlength{\abovecaptionskip}{0.1cm}
\setlength{\belowcaptionskip}{-0.3cm}
	\caption{ Runtimes and model size between our methods and SOTA methods on 30 frames of a test scene, using an Intel(R) Xeon(R) Gold 6230 CPU and an NVIDIA Quadro RTX 8000 GPU. 
}
\centering
{\begin{tabular}{c|c c c }
    \hline		
Methods    & Avg.Time(ms) & Model Size(M)   \\ \hline
ESTRNN & 35 & 2.47 \\
FGST           &  834 & 9.70 \\
LEDNet  & 43 & 7.41 \\
VRT      & 298 & 18.32 \\
RVRT & 442 & 13.57\\ 
Ours                        & 692 & 12.17 \\\hline
	\end{tabular}}
	\label{tab:runtime} 
\end{table}

\subsubsection{Qualitative Analysis.}
Qualitative visual results for Joint Deblurring, Low-light Enhancement and Denoising on our MLBN dataset are shown in \cref{fig:result1,fig:fig1}.
Overall, the results show that our method has two main advantages over other methods. Firstly, the results of our approach have more precise and sharper details, such as the `text board'  in \cref{fig:fig1} and the "Shop logos on the street" in \cref{fig:result1}. It shows that the frames processed with our method are more realistic than others.
Second, blur in real-world videos often exhibits non-uniformity, with intensity varying between severe and mild blurring. \cref{fig:result1} illustrates that our approach outperforms other methods in scenarios characterized by uneven noise-induced blurring under low-light conditions. This holds true for light enhancement, denoising, and deblurring aspects.

\begin{figure*}[htb]
    \setlength{\abovecaptionskip}{1.5pt}
    \centering
    % 第一行
    \begin{subfigure}{.25\textwidth}
        \centering
        \includegraphics[width=0.95\linewidth]{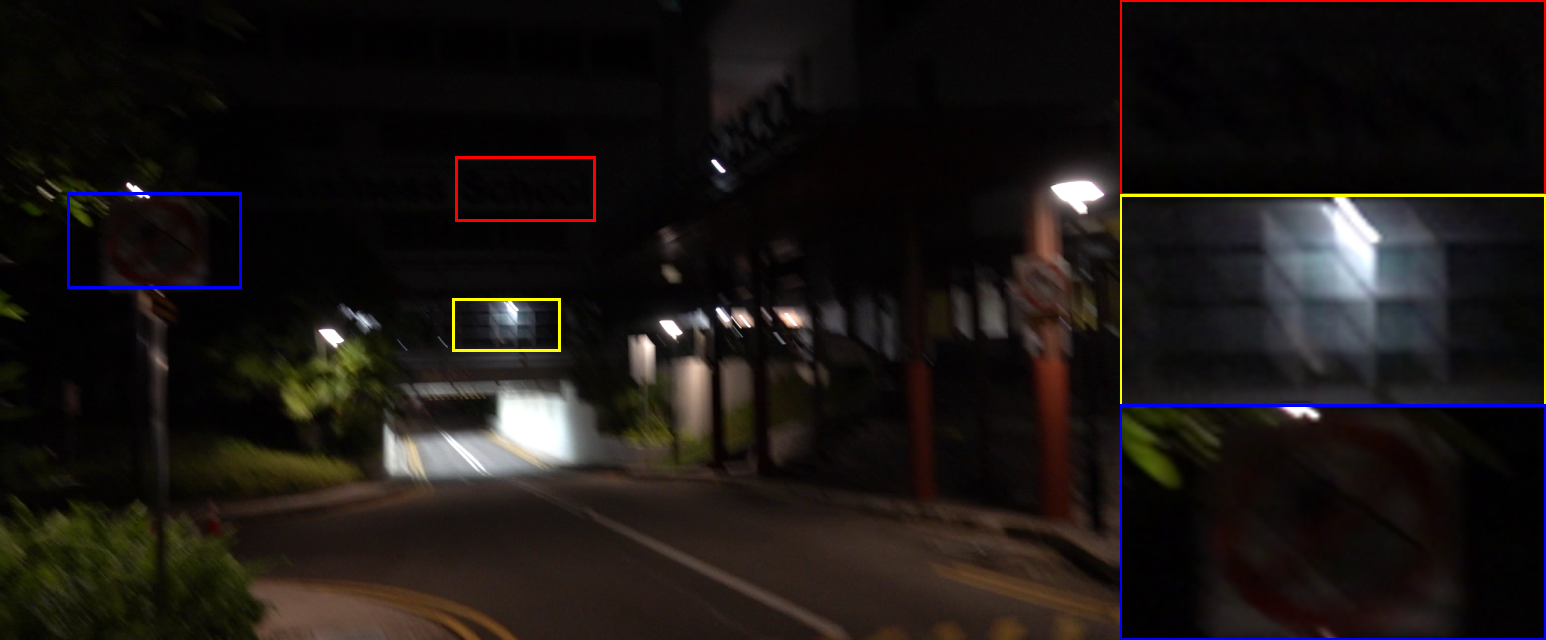}
        \caption{LQ}
        %\label{fig:sub1}
    \end{subfigure}%
    \begin{subfigure}{.25\textwidth}
        \centering
        \includegraphics[width=0.95\linewidth]{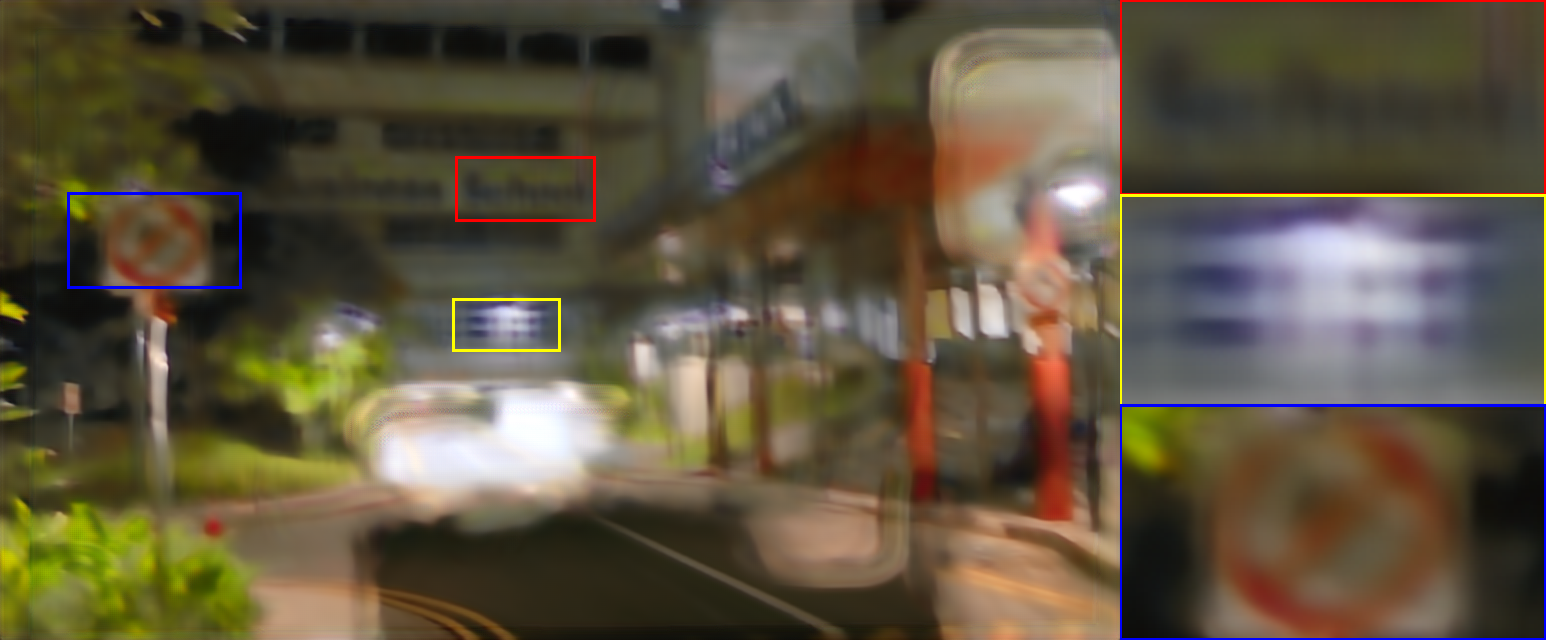}
        \caption{TSP \citeauthor{Pan2020TSP}}
        %\label{fig:sub2}
    \end{subfigure}%
    \begin{subfigure}{.25\textwidth}
        \centering
        \includegraphics[width=0.95\linewidth]{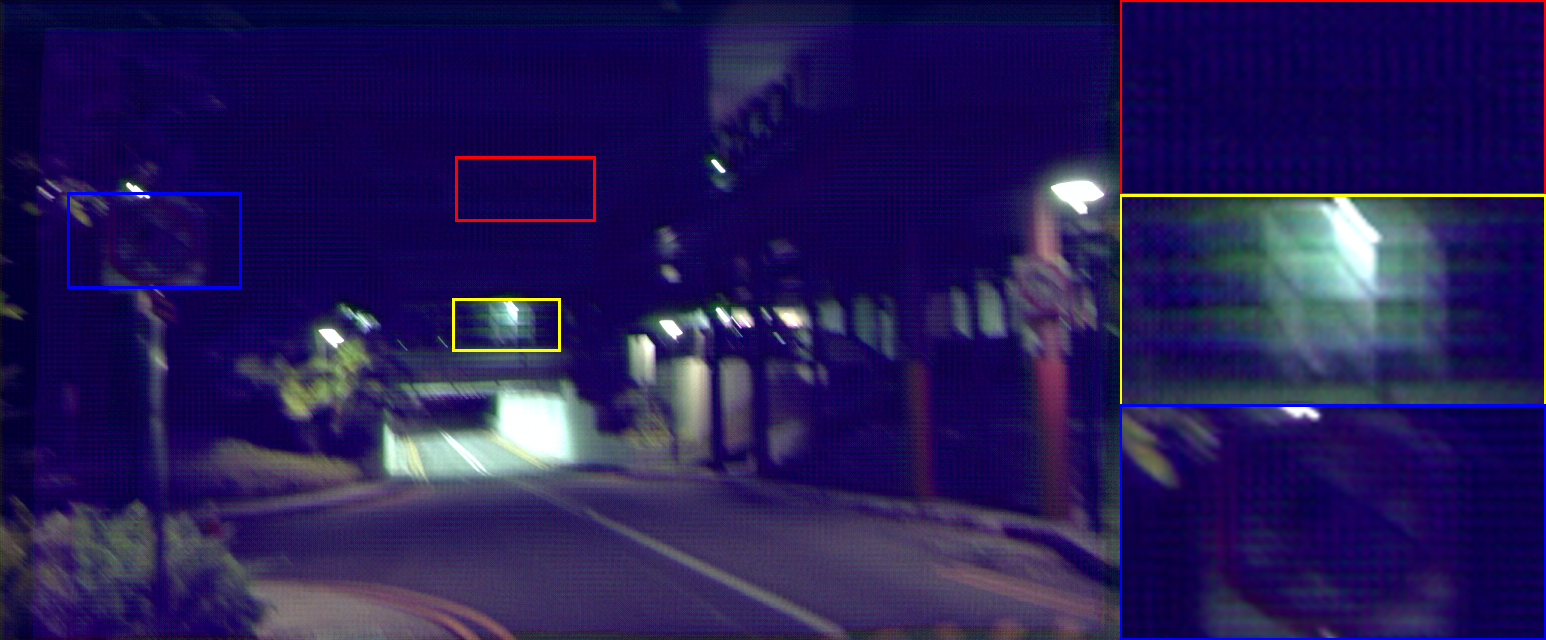}
        \caption{FGST \citeauthor{fgst}}
        %\label{fig:sub3}
    \end{subfigure}%
    \begin{subfigure}{.25\textwidth}
        \centering
        \includegraphics[width=0.95\linewidth]{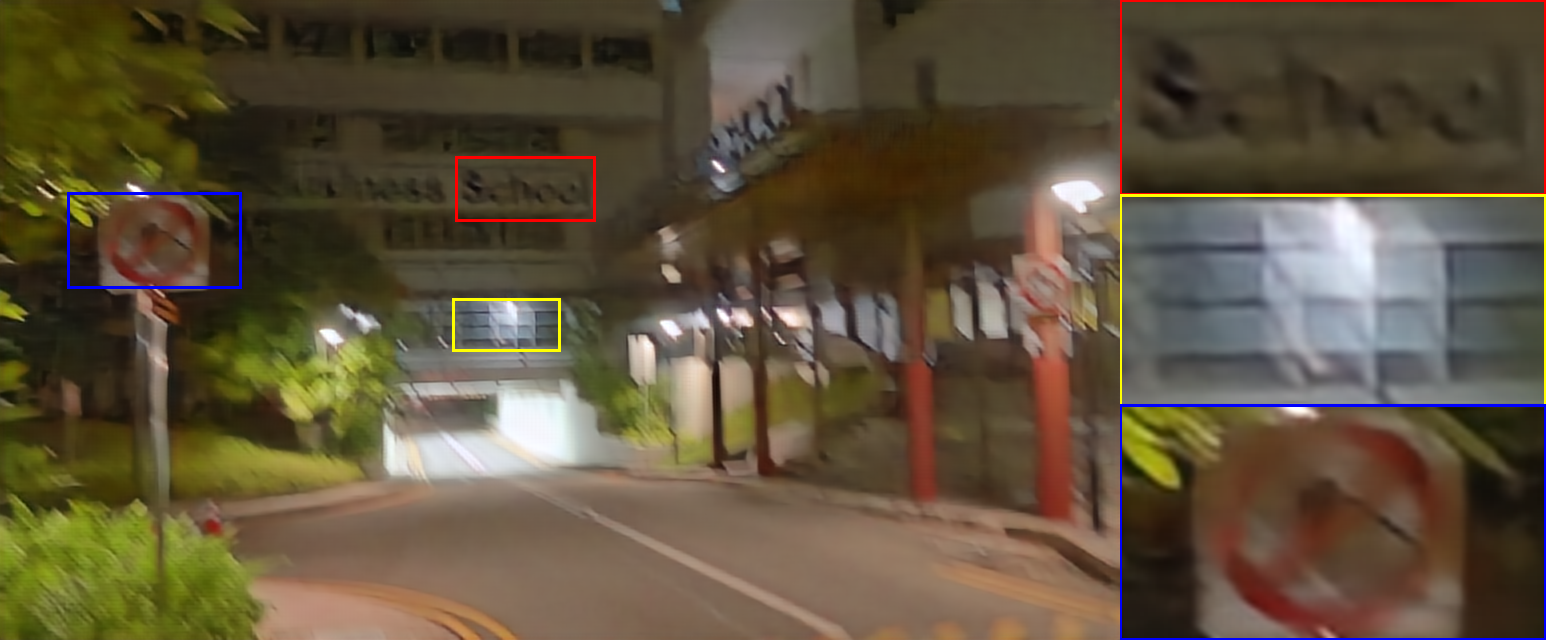}
        \caption{ESTRNN \citeauthor{zhong2020estrnn}}
        %\label{fig:sub4}
    \end{subfigure}%

    % 第二行
    \begin{subfigure}{.25\textwidth}
        \centering
        \includegraphics[width=0.95\linewidth]{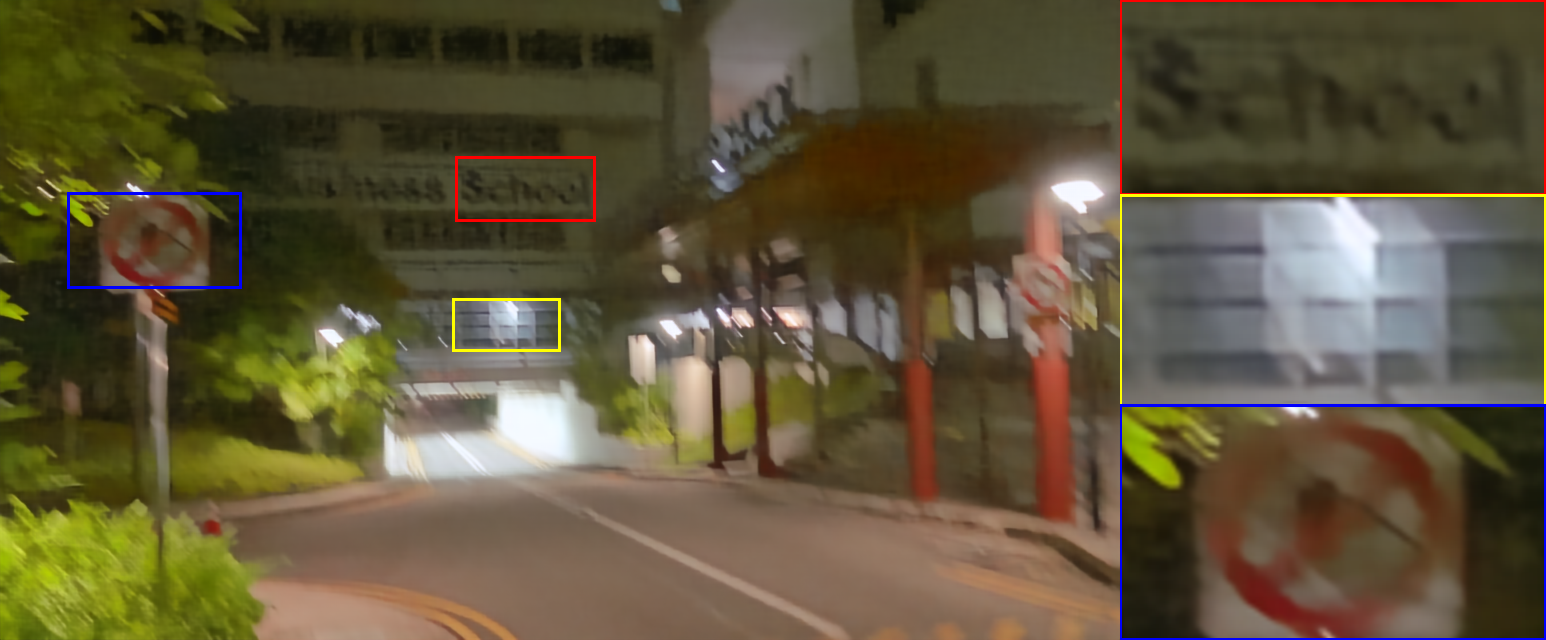}
        \caption{VRT \citeauthor{liang2022vrt}}
        %\label{fig:sub5}
    \end{subfigure}%
    \begin{subfigure}{.25\textwidth}
        \centering
        \includegraphics[width=0.95\linewidth]{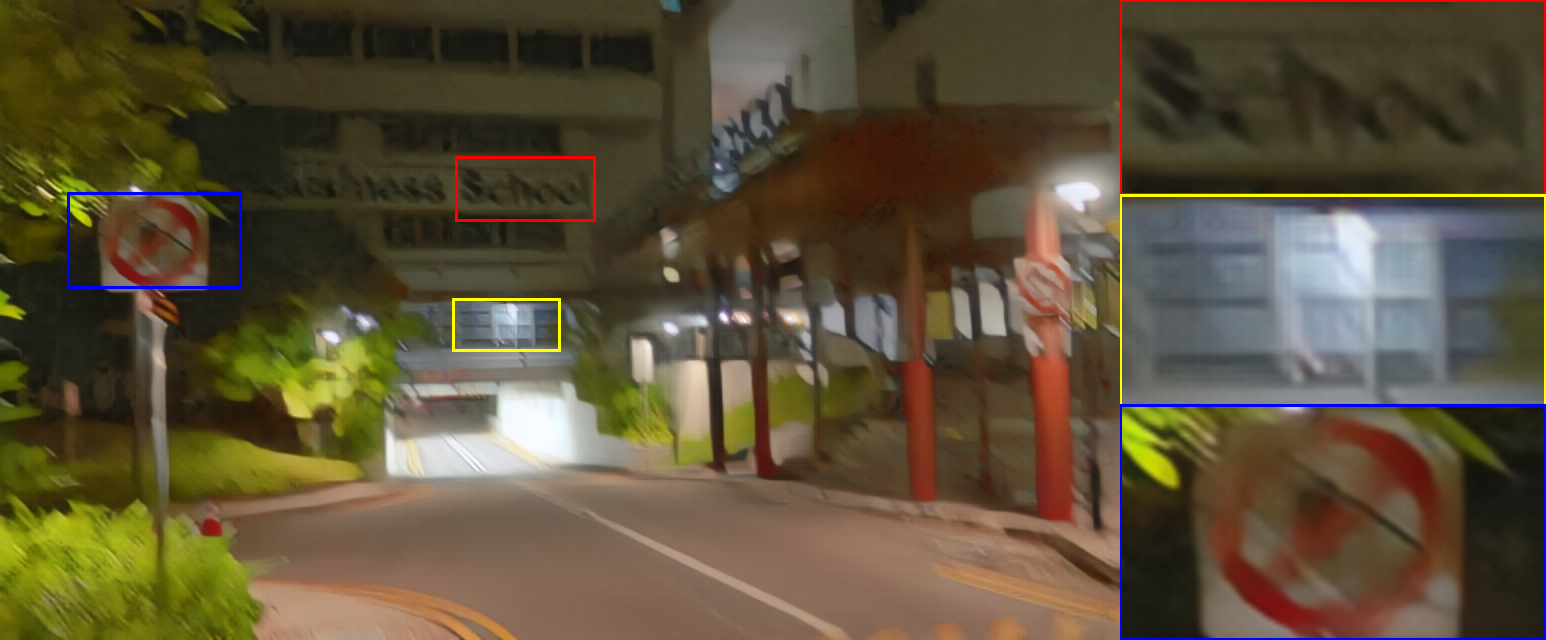}
        \caption{LEDNET \citeauthor{zhou2022lednet}}
        %\label{fig:sub6}
    \end{subfigure}%
    \begin{subfigure}{.25\textwidth}
        \centering
        \includegraphics[width=0.95\linewidth]{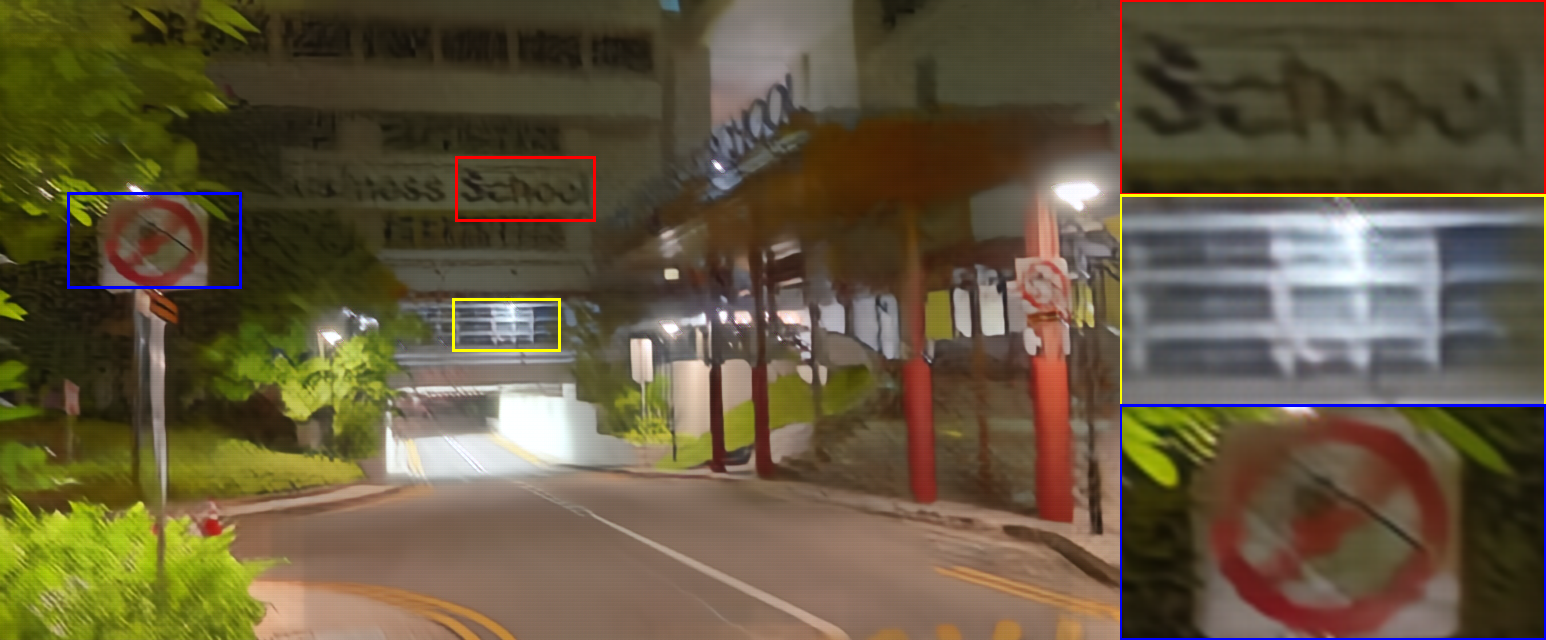}
        \caption{RVRT \citeauthor{liang2022recurrent}}
        %\label{fig:sub7}
    \end{subfigure}%
    \begin{subfigure}{.25\textwidth}
        \centering
        \includegraphics[width=0.95\linewidth]{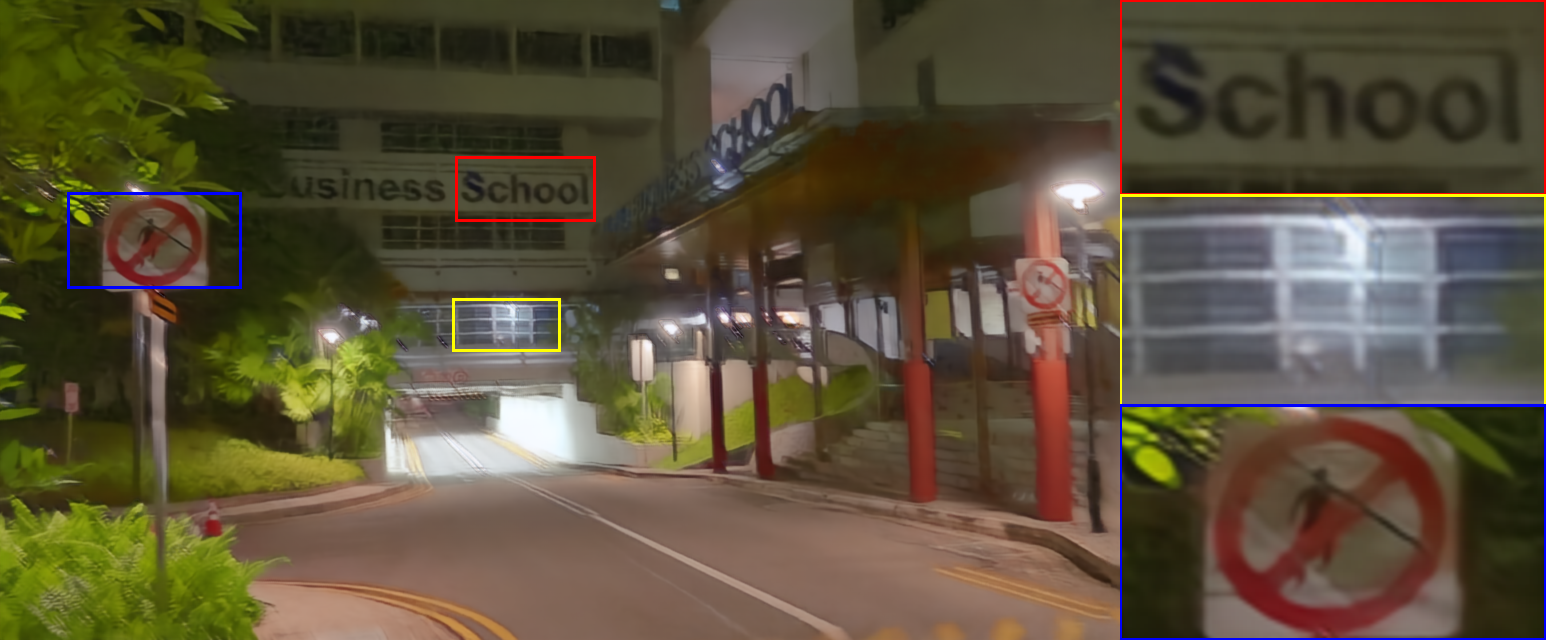}
        \caption{Ours}
        %\label{fig:sub8}
    \end{subfigure}%
    
    \caption{Experiment conducted on the real-lol-blur dataset provided by LEDNet. It can be observed that our method achieves sharper and brighter image restoration.}
    \label{fig:real-lol-blur}
    
\end{figure*}

As for real-world datasets, our model has been tested on the real-lol-blur dataset introduced by LEDNet, and the results are presented in \Cref{fig:real-lol-blur}.

% Ablation study Visual results

\subsection{Compare to three stage concatenation methods}

We employed a three-stage concatenation approach. For the Deblur stages, we utilized highly effective video methods, RVRT. As for the low-light enhancement phase, we employed the video method StableLLVE and image method Zerodce++ separately. And for the denoise stage, RVRT is also the SOTA method for video denoising.The results are in \cref{tab:psnr2,tab:psnr3}.
We also compared different concatenation orders of
Denoise, Deblur, and Lowlight Enhancement in \cref{tab:psnr3} by using RVRT-Deblur,StableLLVE-LowlightEnhancement and RVRT-Denoise.
On our dataset, we performed fine-tuning for both the RVRT model used for Deblur and Denoise tasks and the StableLLVE model utilized for low-light enhancement.

Within the progressive concatenation framework, the three distinct Ground Truths are derived from corresponding stages of different synthetic datasets, with the sequence of restoration tasks in the model mirroring the data generation sequence.

% compare concanate methods
% 从表格\ref{tab:psnr2,tab:psnr3}中可以看出我们的方法都远超过finetune之后的拼接方法，在PSNR和SSIM上大幅度领先。并且我们验证了不同顺序的拼接方法会带来不同程度的恢复效果，PSNR上下相差最大1dB,SSIM相差最大0.05

From the data presented in Tables \ref{tab:psnr2} and \ref{tab:psnr3}, it is evident that our approach significantly outperforms the post-finetuning concatenation methods, achieving substantial leads in both PSNR and SSIM metrics. Additionally, we have verified that the order of concatenation in the methodology can lead to varying degrees of restoration effectiveness. The maximum difference observed in PSNR was approximately 1dB, and in SSIM, the maximum variation was up to 0.05.

In the concatenation experiments section of our comparative analysis, we indirectly demonstrated that there are discernible differences between varying sequences;The sequence we adopted in our model is aligned with the synthetic procedure of the dataset, which involves progressive Ground Truth data. This sequentiality is deliberate since video restoration serves as the inverse process to the methodical degradation encoded in the dataset generation.

\begin{table}[htbp]
\setlength{\abovecaptionskip}{0.1cm}
\setlength{\belowcaptionskip}{-0.3cm}
	\caption{Quantitative results compared with concatenation methods}
	\centering
	{\begin{tabular}{c|c c }
			\hline		
			Deblur-LightEnhancement-Denoise    & PSNR  & SSIM     \\ \hline
			RVRT-Zerodce++-RVRT   & 14.90 & 0.2368    \\
            RVRT-StableLLVE-RVRT   & 16.75 &  0.3152   \\ \hline
			Ours       & {\bf 25.45} & {\bf 0.8083}   \\ \hline
	\end{tabular}}
	\label{tab:psnr2} 
\end{table}

\begin{table}[htbp]
\setlength{\abovecaptionskip}{0.1cm}
\setlength{\belowcaptionskip}{-0.3cm}
	\caption{Different concatenation orders of Denoise, Deblur, and Lowlight Enhancement using retrained \textbf{RVRT-Denoise}, \textbf{StableLLVE-LowlightEnhancement} and \textbf{RVRT-Deblur}}
	\centering
	{\begin{tabular}{c|c c }
			\hline		
			Methods    & PSNR$\uparrow$  & SSIM$\uparrow$     \\ \hline
            Deblur-LightEnhancement-Denoise   & 16.75  &  0.3152   \\ 
            Deblur-Denoise-LightEnhancement   & 17.087 &  0.315   \\ 
            LightEnhancement-Deblur-Denoise   & 16.04 &  0.2646   \\ 
            LightEnhancement-Denoise-Deblur   & 16.04 &  0.2649   \\ 
            Denoise-Deblur-LightEnhancement   &  17.002 &  0.3190   \\ 
            Denoise-LightEnhancement-Deblur   & 16.357 &  0.2685   \\ \hline
	\end{tabular}}
	\label{tab:psnr3} 
\end{table}

\subsection{Ablation Study}

We present an ablation study to validate the effectiveness of the main components in VJT. \Cref{tab:ablation} are the quantitative table of the ablation experiment.

\subsubsection{Effectiveness of Multi-tier Architecture and Feature Fusion.}
The multi-tier framework and feature fusions are the foundation and core of our method, and removing them means VRT with channel size(48,60). Experiments show that our results are much higher than the VRT on PSNR and SSIM metrics.

% Ablation Study Table
\begin{table}[htpb]
\setlength{\abovecaptionskip}{0.1cm}
\setlength{\belowcaptionskip}{-0.3cm}
\caption{ Ablation study on the MLBN dataset.}
	\centering
	{\begin{tabular}{c|c c }
			\hline		
			Methods    & PSNR$\uparrow$ & SSIM$\uparrow$     \\ \hline
                w/o multi-tier Architecture  & {23.37} & {0.7430} \\
                w/o Adaptive Weight Scheme  & {24.71} & {0.7831} \\
			Ours       & {\bf 25.45} & {\bf 0.8083}  \\ \hline
	\end{tabular}}
	\label{tab:ablation} 
\end{table}

\subsubsection{Effectiveness of Adaptive Weight Scheme.}
During the training process, we observed that the ratio between the Losses becomes progressively smaller as the training converges. Simultaneously, the ratio of coefficients selected by the Adaptive Weight Scheme also diminishes, which shows that it can automatically adjust the energy levels of multiple loss functions. The traditional grid method of setting coefficients, on the other hand, takes a lot of time to adjust the coefficients of the loss functions and often fails to find the best combination.

\section{Conclusion}
% polish by chatgpt4 using paragraph below 
In this paper, we introduce a multi-tier video transformer (VJT) tailored for joint tasks of video deblurring, low-light enhancement, and denoising.  Our VJT used videos with different degrees of degradation with multi-tier, and the feature fusion between tiers can learn features of the joint task progressively and gain better results. 
Moreover, we utilise adaptive weight loss for faster training, which can also improve model performance.
Furthermore, we proposed a data generation progress and made a new Multi-scene Low-light Blur Noise (MLBN) dataset, which approximates various realistic scenes.
Our network and dataset are both innovative and provide the basis of joint video tasks in the future. Experiments have clearly illustrated the leading performance of the proposed method. More results will be provided in the supplementary part. Dataset and Codes will be public.

\bibliography{aaai24}

\end{document}